\documentclass[lettersize,journal]{IEEEtran}
\usepackage{amsmath,amsfonts}
\usepackage{algorithmic}
\usepackage{algorithm}
\usepackage{array}
\usepackage[caption=false,font=normalsize,labelfont=sf,textfont=sf]{subfig}
\usepackage{textcomp}
\usepackage{stfloats}
\usepackage{url}
\usepackage{verbatim}
\usepackage{graphicx}
\usepackage{hyperref}
\usepackage{cite}
\usepackage{multirow}
\begin{document}

\title{Survey of Multimodal Geospatial Foundation Models: Techniques, Applications, and Challenges}

\author{Liling~Yang,
        Ning~Chen,
        Jun~Yue,
        Yidan~Liu,
		Jiayi~Ma,~\IEEEmembership{Senior~Member,~IEEE},\\
        Pedram Ghamisi,~\IEEEmembership{Senior~Member,~IEEE},
        Antonio Plaza,~\IEEEmembership{Fellow,~IEEE},
        and Leyuan~Fang,~\IEEEmembership{Senior~Member,~IEEE}
        
	\thanks{This work was supported in part by the National Natural Science Foundation of China under Grant 62425109 and Grant U22B2014; in part by the Science and Technology Plan Project Fund of Hunan Province under Grant 2022RSC3064; and in part by the Hunan Provincial Natural Science Foundation of China under Grant 2025JJ40059. \emph{(Liling Yang and Ning Chen contributed equally to this work.) (Corresponding author: Leyuan Fang.)}}
        \thanks{Liling Yang, Yidan Liu and Leyuan Fang are with the School of Artificial Intelligence and Robotics, Hunan University, Changsha 410082, China (e-mail: lilingyang162@gmail.com, yidanliu@hnu.edu.cn; fangleyuan@gmail.com).}%
        
        \thanks{Ning Chen is with the Institute of Remote Sensing and Geographic Information System, Peking University, Beijing 100871, China (e-mail: chenning0115@pku.edu.cn).}
         
        \thanks{Jun Yue is with the School of Automation, Central South University, Changsha 410083, China (e-mail: junyue@csu.edu.cn).}
        
		\thanks{Jiayi Ma is with the Electronic Information School, Wuhan University, Wuhan 430072, China (e-mail: jyma2010@gmail.com).}%

        \thanks{Pedram Ghamis is with Helmholtz-Zentrum Dresden-Rossendorf, Helmholtz Institute Freiberg for Resource Technology, 09599 Freiberg, Germany, and also with the Lancaster Environment Centre, Lancaster University, Lancaster LA1 4YR, U.K. (e-mail: p.ghamisi@gmail.com).}
        
        \thanks{Antonio Plaza is with the Hyperspectral Computing Laboratory, Department of Technology of Computers and Communications, Escuela Politécnica, University of Extremadura, 10003 Cáceres, Spain.}

	\thanks{}}
\markboth{}%
{Shell \MakeLowercase{\textit{et al.}}: A Sample Article Using IEEEtran.cls for IEEE Journals}
\maketitle

\begin{abstract}
Foundation models have transformed natural language processing and computer vision, and their impact is now reshaping remote sensing image analysis. With powerful generalization and transfer learning capabilities, they align naturally with the multimodal, multi-resolution, and multi-temporal characteristics of remote sensing data. To address unique challenges in the field, multimodal geospatial foundation models (GFMs) have emerged as a dedicated research frontier.
This survey delivers a comprehensive review of multimodal GFMs from a modality-driven perspective, covering five core visual and vision-language modalities. We examine how differences in imaging physics and data representation shape interaction design, and we analyze key techniques for alignment, integration, and knowledge transfer to tackle modality heterogeneity, distribution shifts, and semantic gaps. Advances in training paradigms, architectures, and task-specific adaptation strategies are systematically assessed alongside a wealth of emerging benchmarks.
Representative multimodal visual and vision-language GFMs are evaluated across ten downstream tasks, with insights into their architectures, performance, and application scenarios.
Real-world case studies, spanning land cover mapping, agricultural monitoring, disaster response, climate studies, and geospatial intelligence, demonstrate the practical potential of GFMs. Finally, we outline pressing challenges in domain generalization, interpretability, efficiency, and privacy, and chart promising avenues for future research.
\end{abstract}


\section{Introduction}
\IEEEPARstart{F}{oundation} models (FMs) have been defined in multiple ways across research, industry, and policy, reflecting their broad impact.
The original definition from Stanford's Center for Research on Foundation Models describes them as ``any model trained on broad data (generally using self-supervision at scale) that can be adapted to a wide range of downstream tasks'' \cite{bommasani2021opportunities}.
Industrial viewpoints expand this by emphasizing scale, architectural choices (e.g., transformer or diffusion), and multimodal adaptability, as noted by Gartner \cite{gartner2023foundation}, while IBM stresses pretraining on broad unlabeled data and minimal fine-tuning for diverse applications \cite{ibm2021foundation}.
Techopedia further contrasts FMs with narrow artificial intelligence (AI), highlighting massive datasets, parameter counts, and knowledge transfer capabilities \cite{techopedia2023foundation}.
Policy frameworks provide more formal criteria: the United States Executive Order defines FMs as generally self-supervised models trained on broad data with at least tens of billions of parameters and applicability across contexts \cite{useo2023foundation}, whereas the European Union AI Act proposal frames them as AI models ``trained on broad data at scale, designed for generality of output, and adaptable to a wide range of distinctive tasks'' \cite{europeancommission2024ai}.
Complementary analyses, such as those from the Ada Lovelace Institute, emphasize societal implications, emergent capabilities, and regulatory challenges. \cite{adalovelace2022foundation}.
Later discussions in fields like computational science establish formal criteria for generality, scalability, re-usability, and even physical consistency \cite{choi2025computational}.
Overall, these perspectives converge on viewing FMs as large-scale, general-purpose models trained on broad data and adaptable to diverse downstream tasks, while differing in the emphasis placed on scale, modality, or governance.

Here, we define an FM as a large-scale AI model, pretrained on diverse data, that learns general representations and enables adaptation to many downstream tasks across different domains and modalities. The pretraining employs either inherent or constructed supervision signals to efficiently adapt to diverse downstream tasks via lightweight adaptation techniques like fine-tuning. The core idea is to establish a unified feature learning paradigm that provides a generalizable representation framework for multi-task and multi-modal settings.
The development of FMs has progressed from language to vision and then to cross-modal domains. In natural language processing (NLP), BERT \cite{devlin2019bert} captures bidirectional contextual semantics via masked modeling. In computer vision (CV), Vision Transformer (ViT) \cite{dosovitskiy2020image} introduces Transformer architectures to vision tasks, overcoming the constraints of convolution and modeling long-range dependency. The segment-anything model (SAM) \cite{kirillov2023segment} enables universal image segmentation. In cross-modal domains, contrastive language-image pre-training (CLIP) \cite{radford2021learning} aligns image-text representations via contrastive learning, while ChatGPT \cite{achiam2023gpt} demonstrates the potential of large language models for intelligent agents.

Despite the remarkable breakthroughs of FMs in NLP and CV \cite{awais2025foundation}, directly applying these general-purpose models to remote sensing (RS) image analysis remains challenging due to the domain gaps.
RS data cover a wide range of spatial resolutions, from tens of meters to sub-meter scales, capturing large-scale land cover and fine-grained textures.
They also contain more spectral dimensions than RGB natural images, such as multispectral (MS) and hyperspectral (HS) data, which are information-rich yet introduce noise and redundancy.
In addition, many tasks rely on temporal observations, e.g., land-use change and disaster monitoring, which require to learn spatiotemporal representations.
RS data involves a wide variety of tasks, like scene-, regional-, and pixel-level tasks. These factors limit the direct adaptation of FMs pretrained on natural images and highlight specific designs.

Recent studies have proposed GFMs to address RS-specific characteristics. For example, RingMo \cite{sun2023ringmo} employs a generative framework to detect dense and small objects in complex scenes, improving imagery interpretation accuracy. ROMA \cite{wang2025roma} introduces a self-supervised auto-regressive framework based on the Mamba architecture, enabling efficient modeling of high-resolution RS data with sparse, multi-scale, and arbitrarily oriented objects. These methods support multiple downstream tasks but lack multimodal support and have limited compatibility.
\begin{figure*}[!t]
	\centering
	\includegraphics[width=\textwidth]{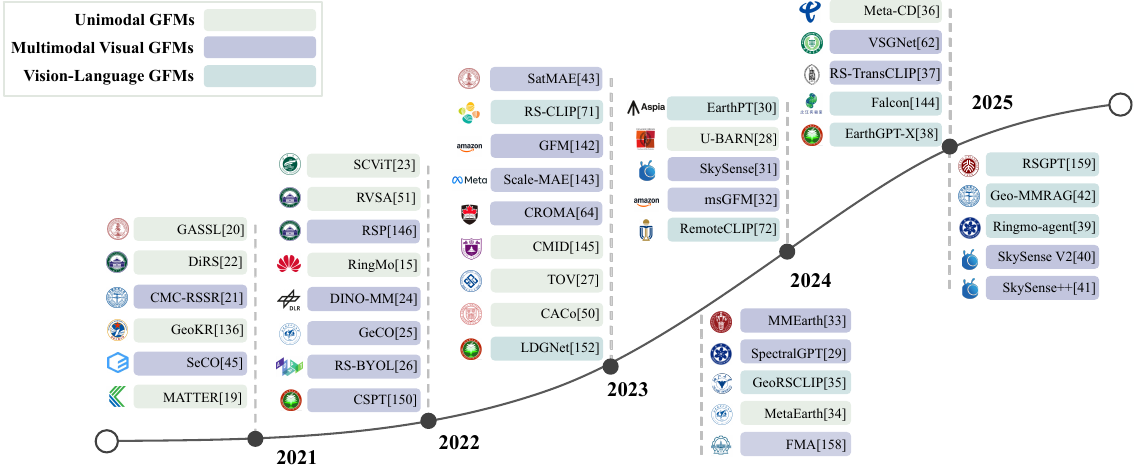}
	\caption{Review of representative geospatial foundation models.}
	\label{figure1}
\end{figure*}
Building on these efforts, the rise of multimodal learning has driven the development of multimodal GFMs. These models integrate complementary information from multiple sensors to enhance robustness and generalization. In land use and land cover classification \cite{dagne2023fusion}, MS and synthetic aperture radar (SAR) provide complementary spectral and structural information, respectively. Combining optical RGB with light detection and ranging (LiDAR) in disaster monitoring \cite{chen2025uav} enables accurate damage assessment with anti-jamming capability. These examples show the potential of multimodal GFMs in complex scenarios.

However, multimodality also brings forth several critical challenges. First, there exists substantial modality heterogeneity. Different modalities exhibit pronounced disparities in imaging mechanisms, viewpoint, spatial and temporal resolution, spectral range, and noise. It is hard to map all modalities into a shared feature space and align spatial-temporal domains.
Second, discrepancies in modality distribution further exacerbate the difficulties. Different types of sensors vary in spatial coverage, revisit frequency, and radiometric response, resulting in data imbalance and sparsity across modalities. These distribution shifts not only hinder stable cross-modal representation learning, but also introduce biases that favor dominant modalities, ultimately limiting the generalization ability of models in long-tailed scenarios.
Finally, there exist intrinsic semantic gaps among modalities. Such gaps happen not only between imaging modalities, but also across vision and language modalities. RS scenes often lack salient foreground objects and exhibit high background redundancy, amplifying semantic misalignment during cross-modal fusion. Direct integration without resolving such latent discrepancies may lead to feature redundancy and suboptimal performance. Taken together, these challenges indicate that multimodal GFMs requires to contend not only with the inherent complexity of RS data but also with the imperative to design robust and generalizable cross-modal modeling frameworks.

As illustrated in Fig.~\ref{figure1}, the development of GFMs can be categorized into three stages: unimodal GFMs, multimodal visual geospatial foundation models (MV-GFMs), and vision-language geospatial foundatio models (VL-GFMs).
Early studies mainly adopted convolutional neural network (CNN) backbones \cite{akiva2022self}, either pretrained on single-source RS datasets \cite{ayush2021geography} or transferred from natural images \cite{stojnic2021self} to address data scarcity. MillionAID \cite{long2021creating}, the first million-scale RS dataset, marked a milestone enabling large-scale pretraining.
By 2022, transformer architectures \cite{lv2022scvit} gradually supplanted CNNs as the dominant backbone, and the significance of multimodal tasks became increasingly recognized \cite{wang2022self, li2022geographical, jain2022self}.
In 2023, VL-GFMs emerged, accompanied by more diverse pretraining strategies, such as self-distillation and continual pretraining. Research focus expanded from dataset scaling to multi-sensor and cross-domain learning for complex scenarios \cite{tao2023tov}.
In 2024, architectures diversified further, incorporating hybrid backbones \cite{dumeur2024self}, generative paradigms \cite{hong2024spectralgpt}, and autoregressive designs \cite{smith2023earthpt}. Multimodal GFMs began structured modeling of modality-specific differences \cite{guo2024skysense, han2024bridging, nedungadi2024mmearth}, while large-scale datasets and benchmarks \cite{yu2024metaearth, nedungadi2024mmearth, zhang2024rs5m} facilitated systematic data construction.
By 2025, cross-task generalization evolved from few-shot to zero-shot transfer \cite{gao2025combining, el2025enhancing}, enhancing adaptability to open-world scenarios. VL-GFMs further advanced to support heterogeneous multi-source RS data \cite{zhang2025earthgpt, hu2025ringmo, zhang2025skysense}.
Overall, the evolution of GFMs has progressed from multimodal compatibility toward cross-task generalization, branching into application-driven multimodal GFMs \cite{wu2025semantic, chen2025vision} with real-world geospatial understanding and language-driven universal agents \cite{hu2025ringmo} with open reasoning.

Unlike most existing surveys, which confine multimodal representation learning to vision-language settings, this work adopts a modality-based perspective for diverse RS data. We analyze how the intrinsic properties of different modalities fundamentally shape their interactions within multimodal systems, moving beyond treating multimodality as a mere addition to task evaluation. We further explore how these modality-specific characteristics guide the design principles, technical innovations, and evaluation of GFMs. Distinct from task-oriented surveys, our approach centers on the modalities themselves and offers a principled understanding of multimodal integration from the perspective of modality characteristics. We provide a systematic reference for future advancements in multimodal GFMs. Thus, our contributions are threefold:

\begin{itemize}
\item{We review the innovations for multimodal GFMs in training paradigms, architectures, and adaptation strategies, outlining their evolution and revealing how to balance modality-specific properties with unified cross-modal representations.}
\item{We propose a modality-based evaluation framework for GFMs that encompasses six representative modalities and their typical combinations, together with the data properties, benchmark, and tasks, and analyze model performance and evolving trends.}
\item{We summarize the key challenges that current multimodal GFMs aim to address and the practical value in real-world applications, providing a critical discussion on promising research directions for future exploration.}
\end{itemize}

\section{Technologies for Multimodal Geospatial Foundation Models}
In this section, we examine the fundamentals of general GFMs and discuss how they technically address multimodal RS data.
\begin{figure}[!h]
	\centering
	\includegraphics[width=\columnwidth]{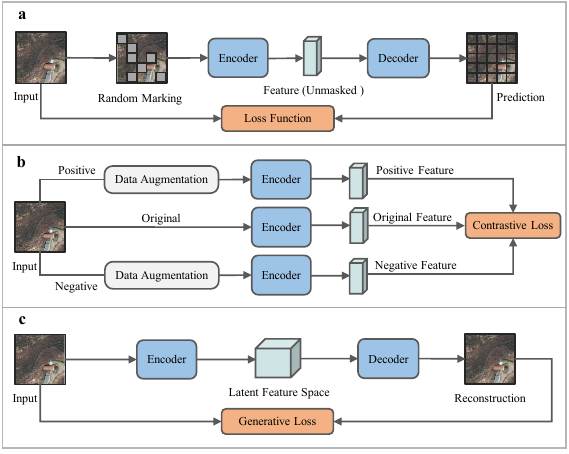}
	\caption{Graphical illustration of training paradigms. (a) Masked modeling, (b) Contrastive learning, (c) Generative learning.}
	\label{figure2}
\end{figure}
\subsection{General Geospatial Foundation Models}
To systematically analyze the core technologies of GFMs, this section discusses three key aspects: training paradigms, network architectures, and adaptation strategies.
\subsubsection{Training Paradigms}
As shown in Fig.~\ref{figure2}, training paradigms enable universal representation in GFMs to learn from unlabeled and multimodal data.

\textbf{Masked modeling (MM)} is a self-supervised paradigm that learns general representations by randomly masking and reconstructing inputs. It encompasses masked language modeling (MLM) and masked image modeling (MIM). BERT \cite{devlin2019bert} pioneered MLM by bidirectionally predicting masked tokens, while MIM extends this idea to image patches to accommodate image continuity and information density. For dense and small targets, RingMo \cite{sun2023ringmo} employs MIM to preserve critical details, learning generalizable representations through reconstruction. SatMAE \cite{cong2022satmae} further validates MAE-based GFMs by introducing temporal embeddings and spectrally aware block masking. MM enables strong cross-modal adaptation and task generalization with limited annotations but remains sensitive to masking ratios across modalities.

\textbf{Contrastive learning (CL)} constructs discriminative representations by maximizing similarity and dissimilarity between positive-negative pairs \cite{chen2020simple}. Leveraging spatiotemporal correlations, it supports cross-modal alignment and temporal change detection. SeCo \cite{manas2021seasonal} learns transferable features via spatial-temporal invariance, mapping features into a shared embedding space. CLIP, a representative CL framework, is widely used in image-text FMs. In visual FMs, CL aligns intra-image features, whereas image-text FMs require semantic alignment across modalities. Overall, CL facilitates multimodal alignment and temporal modeling under limited supervision but depends heavily on the design of positive-negative pairs.

\textbf{Generative learning (GL)} models data distributions to generate new samples \cite{khan2024survey}, effectively capturing rich spatial structures and hierarchical semantics. 
Using typical approaches like auto-regressive and diffusion-based, GL has shown promise in RS domains.
SMLFR \cite{dong2024generative} introduces the first generative convolutional GFM by sparse masking and low-frequency reconstruction to suppress noise and recover land-cover features.
MetaEarth \cite{yu2024metaearth} introduces a resolution-guided self-cascading framework with a novel noise-sampling strategy, facilitating seamless multi-resolution synthesis of RS data.
GL enhances unsupervised reconstruction of complex spatial-semantic structures but requires carefully designed objectives and loss functions to ensure task compatibility.

\begin{figure}[!h]
	\centering
	\includegraphics[width=\columnwidth]{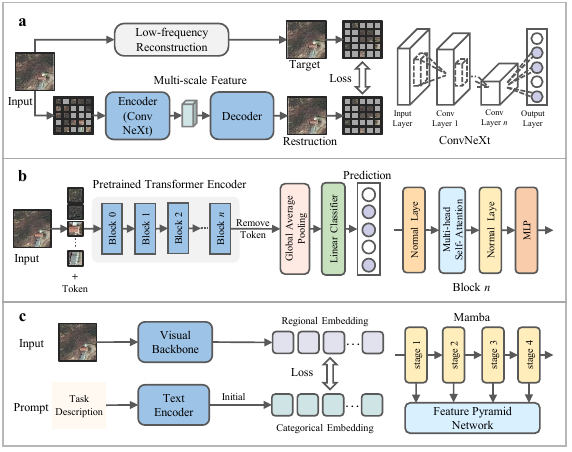}
	\caption{Model architectures across backbones. (a) CNN, (b) Transformer, (c) Hybrid model.}
	\label{figure3}
\end{figure}

\subsubsection{Network Architectures} 
The network architecture directly governs the granularity of feature extraction, cross-modal alignment, and temporal modeling, as analyzed for different backbones in Fig.~\ref{figure3}.

\textbf{Convolutional neural networks (CNNs)} exploit local connectivity and hierarchical representation \cite{pinaya2020convolutional}, enabling multi-scale analysis with fine boundary capture and noise-robust global abstraction. InternImage \cite{wang2023internimage} replaces standard convolutions with deformable operators for adaptive spatial aggregation. Residual network (ResNet)-based methods such as CACo \cite{mall2023change} compute hierarchical feature differences and decode long-term changes via U-Net. Yet, CNNs require extra modules for multimodal generalization, prompting the shift toward complex designs.

\textbf{Transformers} use self-attention to model long-range and cross-modal dependencies \cite{dosovitskiy2020image}, as in ViT and Swin Transformer. They capture global correlations and multi-scale semantics. SatMAE \cite{cong2022satmae} applies temporal-spectral positional encodings, masking both dimensions to disentangle spatio-temporal-spectral features. However, their high computational and parametric cost limits edge deployment. RVSA \cite{wang2023advancing} alleviates this by adopting rotation-variable window attention for large-scale, multi-angle data.

\textbf{Hybrid models} integrate CNNs, Transformers, state space models (SSMs), or other architectures within a unified framework to exploit their complementary strengths. Changen2 \cite{zheng2025changen2} adopts a ViT-based autoencoder and processes intermediate features via a variable-resolution diffusion transformer to generate time series along with semantic and change labels from unlabeled single-temporal images. As an SSM extension, Mamba alleviates the challenge of long-range feature extraction \cite{feng2025hybrid}. DynamicVis \cite{chen2025dynamicvis} couples a CNN-based feature pyramid network with an SSM for efficient, cross-resolution, region-aware semantic representation. By balancing the merits of different architectures, hybrid models represent a promising direction, with future efforts likely emphasizing lightweight designs optimized for edge deployment.

\subsubsection{Task-specific Adaptation Strategies}
Given the diversity of scenes, modalities, and limited annotations, it is crucial to unlock the potential of general GFMs. We summarize three main adaptation strategies.
\begin{figure*}[!h]
	\centering
	\includegraphics[width=\textwidth]{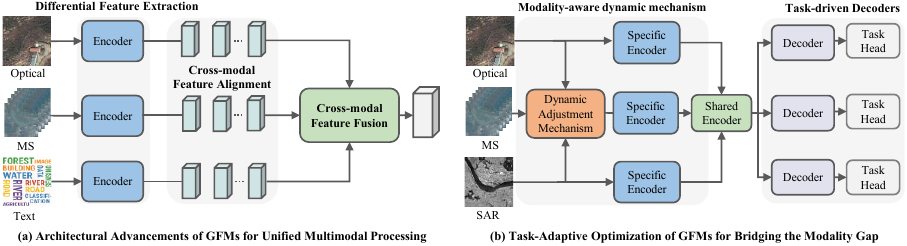}
	\caption{Network architectures and task-specific adaptation strategies for multimodal geospatial foundation models.}
	\label{figure4}
\end{figure*}

\textbf{Fine-tuning} updates part or all of a pre-trained model's parameters using task-specific data \cite{zhu2025layerlink}, transferring generic representations from large-scale pre-training to downstream tasks with limited labels. Parameter-efficient variants such as LoRA, Adapters, and Prompt-tuning enable multi-task adaptation with minimal tuning. SatlasPretrain \cite{bastani2023satlaspretrain} employs a Swin Transformer encoder with seven task-specific heads, yielding a 7.1\% average gain. UPetu \cite{dong2024upetu} applies parameter-efficient fine-tuning (PEFT) for dense prediction, inserting prompts into multi-scale features to generalize with few updated parameters.

\textbf{Few-shot learning} enables competitive performance with scarce annotations \cite{song2023comprehensive}, leveraging prior knowledge for rapid adaptation to novel categories or tasks. This paradigm is particularly useful for GFMs, where labeling is costly and scene diversity is high. FoMA \cite{gao2024enrich} integrates zero-shot segmentation with few-shot learning, balancing base and novel classes through knowledge distillation and expert fusion. Besides, meta-learning, following the “learning-to-learn” principle, equips models with fast task adaptation. MetaPEFT \cite{tian2025meta} extends PEFT via joint optimization of structural parameters, improving generalization across diverse downstream tasks.

\textbf{Prompt learning (PL)} adapts pre-trained models by injecting lightweight, task-specific prompts into inputs or intermediate layers, without modifying model weights. Unlike full fine-tuning, which risks overwriting pre-trained knowledge \cite{khattak2023maple}, prompt learning enables efficient and generalizable adaptation. SAM introduces prompts as flexible conditioning signals for segmentation and recognition, while VSGNet \cite{pan2025vision} leverages SAM to extract morphological prompts guiding RGB semantic segmentation. In vision-language FMs, task-oriented prompts enhance transfer; for instance, Liu \textit{et al.} \cite{liu2023decoupling} design a decoupled multi-prompt scheme integrating pre-trained large language models (LLMs) for change description.

\subsection{Multimodal Geospatial Foundation Models}
Our framework divides multimodal GFMs into two categories: multimodal visual GFM (which focuses on multimodal visual modalities by complementary semantic details, structural patterns, and spatiotemporal continuity), and vision-language GFM (which connects perception to cognition by open-vocabulary understanding, instruction following, and reasoning capabilities).
\subsubsection{Training Paradigms}
Multimodal GFMs contain two core training paradigms, contrastive learning and generative learning, from learning within-modality representations toward establishing a shared semantic space where heterogeneous modalities become comparable and interchangeable.

\textbf{Contrastive learning (CL)} aligns multimodal data by minimizing intra-domain distances while maximizing inter-domain ones, with its effectiveness hinging on the construction of positive-negative sample pairs. In multimodal VM-GFMs, the primary challenge arises from low-level inconsistencies caused by differing imaging principles, which often lead to feature mismatches that degrade alignment. To address this, some approaches decompose features, modeling low-level physical cues separately while aligning high-level semantics. For example, CROMA \cite{fuller2023croma} inputs co-registered MS and SAR data to form sample pairs and enhances features through cross-modal attention.
While traditional CL typically operates at the image level and struggles to balance global and local information, multi-granularity methods construct pairs across scales, channels, or regions to achieve joint alignment. SkySense \cite{guo2024skysense} implements this through pixel-, object-, and image-level losses, incorporating spatiotemporal positional encoding to synchronize feature alignment across modalities.

For VL-GFMs, the challenge lies in mapping dense pixel arrays to discrete linguistic symbols. CL establishes direct pixel-to-semantics correspondence. CLIP \cite{radford2021learning} first unified images and texts in a shared embedding space, enabling multimodal interaction. Following this, GeoRSCLIP \cite{zhang2024rs5m} bridges natural and geospatial understanding without altering architecture, while ChangeCLIP \cite{dong2024changeclip} enhances bitemporal change reasoning via visual-language interaction. These advances underscore CL as the cornerstone of multimodal representation learning in VL-GFM, capturing domain-specific visual nuances and cross-modal semantics crucial for complex reasoning.

\textbf{Generative learning (GL)} builds latent dependencies and semantic consistency across modalities by modeling their joint data distribution. It bridges modality gaps through cross-modal synthesis and shared representation generation. Unlike CL, which emphasizes discriminative alignment, GL learns from regional structures but may overfit texture noise if reconstruction dominates. To overcome this, hierarchical generative frameworks introduce multi-level priors and semantic-space interactions for coarse-to-fine alignment. DiffusionSat \cite{khanna2024diffusionsat} extends diffusion models with temporal modeling and multi-scale hierarchies for time-series prediction, while LHRS-Bot \cite{muhtar2024lhrs} combines vision-language alignment with curriculum learning for progressive multimodal fusion.

For VL-GFMs, autoregressive models such as LLaVA and GPT employ instruction tuning for multimodal dialogue. H2RSVLM \cite{pang2024h2rsvlm} extends LLaVA by integrating multi-level visual representations with linguistic reasoning, enhancing multi-task generalization and factual consistency. SpectralGPT \cite{hong2024spectralgpt} fuses spatial-spectral information via 3D masking and multi-objective reconstruction within a GPT-based architecture for robust self-supervised learning. Diffusion-based models such as MMM-RS \cite{wang2024mmm} further improve adaptation to complex RS scenes through generative diffusion backbones. While auto-regressive models risk error accumulation and reduced spatial coherence, diffusion-based approaches demand carefully designed conditioning for cross-modal alignment.
\begin{table*}[t]
	\setlength{\tabcolsep}{3pt}
	\centering
	\caption{Technical Overview of Multimodal Geospatial Foundation Models. \\`-' Indicates Insufficient Information to Clearly Define the Method or the Absence of Modifications.}
	\label{tab1}
	\begin{tabular}{llllcccccccc}
		\hline
		\multirow{2}{*}{Methods}  & \multirow{2}{*}{Years} & \multirow{2}{*}{Backbone} & \multirow{2}{*}{\begin{tabular}[c]{@{}l@{}}Modality\\ (Optical+)\end{tabular}}            & \multicolumn{2}{c}{Training Paradigm} & \multicolumn{3}{c}{Network Architecture}                                                                                                                                              & \multicolumn{3}{c}{Adaptation Strategy}                                                                                                                   \\ \cline{5-12} 
                                                   &      &                                                           &                                                            & CL                   & GL             & \begin{tabular}[c]{@{}c@{}}Feature\\ Extraction\end{tabular} & \begin{tabular}[c]{@{}c@{}}Feature\\ Alignment\end{tabular} & \begin{tabular}[c]{@{}c@{}}Feature\\ Fusion\end{tabular} & \begin{tabular}[c]{@{}c@{}}Task-driven\\ Decoders\end{tabular} & \begin{tabular}[c]{@{}c@{}}Modality-aware\\ Dynamic Mechanism\end{tabular} & PL          \\ \hline
		CROMA \cite{fuller2023croma}               & 2023 & ViT-B                                                     & MS+SAR                                                     & \checkmark           &                & \checkmark                                                   & \checkmark                                                  &                                                          & \multicolumn{3}{c}{-}                                                                                                                                     \\
		Skysense \cite{guo2024skysense}            & 2024 & Swin-H                                                    & MS+SAR                                                     & \checkmark           &                & \checkmark                                                   & \checkmark                                                  & \checkmark                                               & \multicolumn{3}{c}{-}                                                                                                                                     \\
		DiffusionSat \cite{khanna2024diffusionsat} & 2024 & DiffusionSat                                              & VL                                                         &                      & \checkmark     & \checkmark                                                   &                                                             &                                                          & \multicolumn{3}{c}{-}                                                                                                                                     \\
		LHRS-Bot \cite{muhtar2024lhrs}             & 2024 & \begin{tabular}[c]{@{}l@{}}ViT-L \&\\ LLaMA2\end{tabular} & VL                                                         &                      & \checkmark     & \checkmark                                                   & \checkmark                                                  &                                                          & \multicolumn{3}{c}{-}                                                                                                                                     \\
		LAE-DINO \cite{pan2025locate}              & 2025 & Swin-T                                                    & VL                                                         & \checkmark           &                & \checkmark                                                   & \checkmark                                                  &                                                          &                                                                &                                                                            & \checkmark  \\
		RS-CLIP \cite{li2023rs}                    & 2023 & ResNet/ViT                                                & VL                                                         & \checkmark           &                & \checkmark                                                   &                                                             &                                                          & \multicolumn{3}{c}{-}                                                                                                                                     \\
		RemoteCLIP \cite{liu2024remoteclip}        & 2024 & ViT-B                                                     & VL                                                         & \checkmark           &                & \checkmark                                                   & \checkmark                                                  &                                                          & \multicolumn{3}{c}{-}                                                                                                                                     \\
		EarthView \cite{velazquez2025earthview}    & 2025 & ViT-S                                                     & \begin{tabular}[c]{@{}l@{}}MS+HS\\ +SAR+LiDAR\end{tabular} & \checkmark           &                & \checkmark                                                   & \checkmark                                                  &                                                          & \multicolumn{3}{c}{-}                                                                                                                                     \\
		DOFA \cite{xiong2024neural}                & 2024 & ViT-B                                                     & MS+HS+SAR                                                  & \checkmark           &                & \checkmark                                                   &                                                             &                                                          &                                                                & \checkmark                                                                &              \\
		FlexiMo \cite{li2025fleximo}               & 2025 & ViT-B                                                     & MS+SAR                                                     & \checkmark           &                &                                                              & \checkmark                                                  &                                                          & \multicolumn{3}{c}{-}                                                                                                                                     \\
		Ringmo-SAM \cite{yan2023ringmo}            & 2023 & ViT-L                                                     & SAR                                                        & \multicolumn{2}{c}{-}                 & \checkmark                                                   & \checkmark                                                  &                                                          & \checkmark                                                     &                                                                           & \checkmark   \\
		VSGNet \cite{pan2025vision}                & 2025 & ViT-G                                                     & LiDAR                                                      &                      & \checkmark     &                                                              &                                                             & \checkmark                                               & \multicolumn{3}{c}{-}                                                                                                                                     \\
		SkySense V2 \cite{zhang2025skysense}       & 2025 & SwinV2                                                    & MS+SAR                                                     & \checkmark           &                & \checkmark                                                   & \checkmark                                                  & \checkmark                                               &                                                                &                                                                           & \checkmark   \\
		msGFM \cite{han2024bridging}               & 2024 & SwinV2                                                    & SAR+LiDAR                                                  & \checkmark           &                & \checkmark                                                   &                                                             &                                                          & \checkmark                                                     &                                                                           &              \\
		MTP \cite{wang2024mtp}                     & 2024 & Swin-H                                                    & MS+SAR                                                     & \multicolumn{2}{c}{-}                 &                                                              & \checkmark                                                  &                                                          & \checkmark                                                     &                                                                           &              \\
		RingMoE \cite{bi2025ringmoe}               & 2025 & RingMoE                                                   & MS+SAR                                                     & \checkmark           &                & \checkmark                                                   &                                                             & \checkmark                                               &                                                                & \checkmark                                                                &              \\ \hline
		\end{tabular}
\end{table*}
\subsubsection{Network Architectures}
The heterogeneity of multimodal RS data in terms of resolution, spectral channels and noise necessitates architectures capable of handling diverse inputs and integrating complementary features. As shown in Fig.~\ref{figure4}, a generalized multimodal GFM framework revolves around three architecture improvement and adaptation strategies. At this point, it is important to note that multimodal GFMs share these ideas but differ in architectural emphasis; VM-GFMs focus on geometric or physical alignment, whereas VL-GFMs target high-level semantic correspondence.

\textbf{Differential feature extraction} uses modality-specific encoders to capture distinct distributions. EarthView \cite{velazquez2025earthview} tokenizes each modality and integrates temporal, source, and positional embeddings for adaptive decoupling. Yet excessive modality-specific modeling risks semantic ambiguity, while strict global consistency may obscure physical traits. To balance this, shared-specific hybrid representations are introduced. DOFA \cite{xiong2024neural} employs a wavelength-aware dynamic network and shared backbone to jointly learn modality-shared embeddings and band-specific features via distillation, preserving physics and enabling multimodal transfer.
Given the large gaps in feature distribution and granularity between vision and language, VL-GFMs map features into a unified space, modality-specific prediction, or shared coding. RS-CapRet \cite{silva2024large} introduces linear projections for efficient image-text alignment, while EarthGPT \cite{zhang2024earthgpt} uses a ViT-CNN dual backbone with language encoders to project multi-scale features into the shared semantic.
Designing separate networks for each modality along with inefficient parameter utilization, highlighting the need for feature alignment and fusion. 

\textbf{Cross-modal feature alignment} is typically achieved by explicit constraints, cross-modal attention, or channel mapping. FlexiMo \cite{li2025fleximo} uses dynamic resolution-aware embedding recalibration and spectral-band-specific channel adaptation to unify heterogeneous features with physical consistency. RingMo-SAM \cite{yan2023ringmo} employs adapters to retain the generic feature extraction capability while incorporating modal-specific designs such as SAR polarization decomposition. CL has become the main strategy for cross-modal alignment in the field of visual language. In order not to repeat the CLIP introduction with the previous CL training paradigm, we briefly introduce it here. BITA \cite{yang2024bootstrapping} aligns multi-scale frequency-domain features via image-text CL and fuses frozen visual encoders with LLMs by a lightweight Fourier transform. 

\textbf{Cross-modal feature fusion} combines cross-attention, concatenation, or projection to exploit complementary spatial-spectral-semantic information. VSGNet \cite{pan2025vision} integrates multi-scale attention and dual-branch embeddings for structural-semantic fusion, while SkySense V2 \cite{zhang2025skysense} performs dynamic multi-temporal fusion via temporal attention and prompt tokens. KITR \cite{mi2024knowledge} incorporates external knowledge graphs via cross-attention in bootstrapping language-image pre-training (BLIP), improving retrieval accuracy and domain adaptability. In summary, future designs should balance independent modality modeling with unified alignment, fostering deep semantic interactions for cross-task and spatiotemporal generalization.
\begin{figure*}[h]
	\centering
	\includegraphics[width=\textwidth]{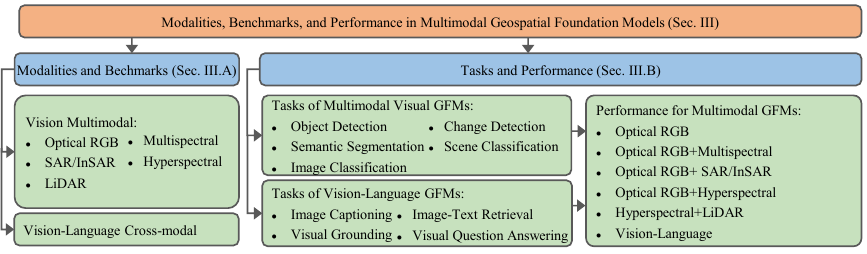}
	\caption{Benchmarks and evaluation framework for multimodal geospatial foundation models.}
	\label{figure5}
\end{figure*}
\subsubsection{Task-specific Adaptation Strategies}
Adaptation strategies for multimodal GFMs address heterogeneous information dimensions by two complementary approaches: 1) structural adaptations employing task-driven decoders and modality-aware dynamic mechanisms to optimize sensor feature utilization, and 2) efficient adaptation strategies utilizing prompt learning and zero-shot learning in VL-GFMs to enable flexible cross-modal reasoning and generalization.

\textbf{Modality-aware dynamic mechanism} employs multi-expert architectures with modality- or task-aware gating. It balances flexibility and computational efficiency by selectively activating shared and specialized experts to generate customized representations on demand. RingMoE \cite{bi2025ringmoe} uses directed expert pruning in the autoencoder, activating shared and modality-specific experts as needed, balancing multi-task flexibility with computational cost. DOFA \cite{xiong2024neural} adopts an autoencoder framework that combines shared and modality-aware learning, producing modality-specific weights on wavelengths to guide the modality-aware dynamic decoding.

\textbf{Task-driven decoders} aim to extract backbone-shared features and tailor them to different task-modality combinations, maximizing the utility of multimodal representations. msGFM \cite{han2024bridging} employs cross-sensor mask modeling and differentiated decoding mechanisms to flexibly reconstruct modalities for downstream tasks. For improving multimodal feature utilization, MTP \cite{wang2024mtp} combines a shared encoder with task-specific decoders under a unified framework, strengthening the encoder customization capacity for diverse task settings.

\textbf{Prompt learning} dynamically directs inference through natural-language instructions, enabling controllable interactions by specifying land-cover categories, semantic granularity, or reasoning chains. Unlike SAM-based GFMs relying on visual prompts, VL-GFMs emphasize linguistic prompts to flexibly indicate spatial and semantic cues. LAE-DINO \cite{pan2025locate} introduces visual-guided textual prompts with dynamic vocabularies for open-vocabulary detection, while RS-TransCLIP \cite{el2025enhancing} fuses category-specific prompts with visual features into frozen LLMs for change description without retraining. Emphasizing controllability and interactivity, prompt learning enhances text-driven tasks with high efficiency.

\textbf{Zero-shot learning} addresses unseen categories or tasks by leveraging semantic similarity within text embeddings, promoting generalization and cross-task transferability. RS-CLIP \cite{li2023rs} combines contrastive vision-language pretraining, pseudo-labeling, and curriculum learning for zero-shot scene classification, while RemoteCLIP \cite{liu2024remoteclip} constructs large-scale image-text datasets via unified label conversion and aligns embeddings through InfoNCE loss. Future work should design robust prompt-engineering frameworks to stabilize outputs and achieve fine-grained vision-language alignment, narrowing the semantic gap in zero-shot inference and advancing VL-GFM toward general-purpose RS interpretation.

Table~\ref{tab1} summarizes advances of unimodal (optical RGB-based) over multimodal GFMs in training strategies, architectures, and adaptation methods.
The above approaches confirm CL's advantages for multimodal integration. Most of them emphasize differentiated processing and feature alignment, though further development of task adaptation strategies remains a promising direction.
\begin{table*}[h]
	\centering
	\caption{Typical Datasets and Downstream Tasks for Geospatial Foundation Models in Optical RGB. }
	\label{tab2}
	\begin{tabular}{llcccc}
		\hline
		Modality                       & Dataset                                                              & Number       & \multicolumn{1}{l}{Class} & Resolution                      & Task                        \\ \hline
		\multirow{17}{*}{Optical RGB}  & DIOR \cite{li2020object}                                       & 23,463       & 20                        & $800\times800$                  & Horizontal Object Detection \\
									   & DIOR-R \cite{cheng2022anchor}                                  & 23,463                     & 20    & $800\times800$                  & Oriented Object Detection   \\
									   & DOTAV1 \cite{xia2018dota}                      & 2,806                      & 15    & $800\times800 \sim20000\times20000$ & Oriented Object Detection   \\
									   & DOTAV2 \cite{ding2022object}                   & 11,268                     & 18    & $800\times800 \sim20000\times20000$ & Oriented Object Detection   \\
									   & FAIR1M \cite{sun2022fair1m}                    & 16,488                     & 37    & $500\times500 \sim1200\times5000$   & Oriented Object Detection   \\
									   & xView \cite{lam2018xview}                      & 846                        & 60    & $2500\times2500 \sim3200\times5000$ & Horizontal Object Detection \\
									   & NWPUVHR \cite{cheng2014multi}               & 800                        & 10    & $500\times500 \sim1100\times1100$ & Horizontal Object Detection \\
									   & LEVIR-CD \cite{chen2020spatial}                & 637 pairs                  & -     & $1024\times1024$                & Change Detection            \\
									   & WHU-CD \cite{ji2019fully}                      & 1 pairs                    & -     & $32507\times15354$              & Change Detection            \\
									   & CDD \cite{lebedev2018change}                   & 16,000 pairs               & -     & $256\times256$                  & Change Detection            \\
									   & iSAID \cite{waqas2019isaid}                    & 2,806                      & 15    & $800 \sim13000$                     & Semantic Segmentation       \\
									   & Potsdam \cite{sherrah2016fully}                & 38                         & 6     & $6000\times6000$                  & Semantic Segmentation       \\
									   & LoveDA \cite{wang2021loveda}                   & 5,987                      & 7     & $1024\times1024$                & Semantic Segmentation       \\
									   & Vaihingen \cite{rottensteiner2012isprs}        & 33                         & 6     & $1996\times1995 \sim3816\times2550$ & Semantic Segmentation       \\
									   & UC Merced (UCM) \cite{yang2010bag}             & 2,100                      & 21    & $256\times256$                  & Scene classification        \\
									   & AID \cite{xia2017aid}                          & 10,000                     & 30    & $600\times600$                  & Scene classification        \\
									   & RESISC45 \cite{cheng2017remote}                & 31,500                     & 45    & $256\times256$                  & Scene classification        \\ \hline
	\end{tabular}
\end{table*}

\section{Modalities, Benchmarks, and Performance in Geospatial Foundation Models}
In Fig.~\ref{figure5}, we outline the evaluation framework for GFMs, systematizing the relationships between inputs, benchmarks, downstream tasks, and performance discussed in this section.
\subsection{Modalities and Benchmarks}
We present a systematic profile of multimodal inputs for GFMs, covering five visual and vision-language data, along with their properties and benchmarks.
\subsubsection{Optical RGB}
With a spectral range of 400 to 700 nm, these data capture surface reflectance in the visible solar spectrum, offering strong interpretability and intuitive information representation. Their spatial detail depends on sensor resolution \cite{verbyla2022satellite}. Owing to these properties, optical RGB data are widely used in land cover classification \cite{temenos2023interpretable}, vegetation analysis \cite{tran2022review}, and disaster management \cite{kucharczyk2021remote}.
As summarized in Table~\ref{tab2}, typical optical RGB datasets contain tens of thousands of samples, often split into training and test sets, serving as both training resources and evaluation benchmarks. For high-resolution data, cropping is commonly adopted for stability and efficiency. For instance, WHU-CD \cite{ji2019fully} (resolution $32507\times15354$) is divided into 6096, 762, and 762 patches of $256\times256$ for training, validation, and testing, respectively, following \cite{zhang2023relation}.

\subsubsection{MultiSpectral (MS)}
These data comprise several narrow spectral bands of the same scene, enabling discrimination by differences in shape, structure, and spectral response. It is particularly useful for land cover classification \cite{yu2022capvit} and precision agriculture \cite{barjaktarovic2024design}.
The datasets in Table~\ref{tab2} reflect typical GFM applications. Some, such as EuroSAT \cite{helber2019eurosat}, provide both RGB and MS data focusing solely on optical inputs, while multimodal datasets like Sen12MS \cite{schmitt2019sen12ms} combine SAR and multispectral data and will be discussed later.
\begin{table*}[]
	\centering
	\caption{Overview of Multimodal Remote Sensing Datasets. \\Notation for Vision-Language Datasets in the `Number': `X/Y' Denotes `X' Images And `Y' Textual Captions. \\(Abbreviations: HTD—Hyperspectral Target Detection, HSR—Hyperspectral Super-Resolution)}
	\label{tab3}
	\begin{tabular}{llccccccc}
	\hline
	Modality       & Dataset         & Number           & Class           & Band         & Resolution         & Task                                                                                     & Multimodal                                 \\ \hline
	\multirow{6}{*}{Multispectral}  & OSCD \cite{daudt2018urban}                     & 24 pairs             & -     &   13  & $600\times600$                 & CD       & Optical       \\
									& Dyna.-S2 \cite{toker2022dynamicearthnet}       & 54,750               & 7     &   12  & $1024\times1024$               & CD, Seg  & Optical       \\
									& Dyna.-Pla \cite{toker2022dynamicearthnet}      & 54,750               & 7     &   12  &$1024\times1024$                & Seg      & Optical       \\
									& GID-15 \cite{tong2020land}                     & 10                   & 15    &   5   &$7200\times6800$                & Seg      & -       \\
									& EuroSAT \cite{helber2019eurosat}               & 27,000               & 10    &   13  &$64\times64$                    & SC       & Optical       \\
									& BigEarthNet (BEN) \cite{sumbul2019bigearthnet} & 590,326              & 43    &   3   &$20\times20 \sim120\times120$   & SC       & -      \\
									& fMow-s2 \cite{cong2022satmae}                  & 132,716              & 63    &  4, 8 &$256\times256$                  & SC       & Optical        \\ \hline 
	\multirow{5}{*}{SAR}
							  & SSDD \cite{zhang2021sar}                  & 1160           & 1 & -    & $500\times500$     & OD    & - \\
							  & EuroSAT-SAR \cite{helber2019eurosat}      & 27,000           & 10 & -   & $64\times64$     & SC    & Optical \\
							  & Sen12MS \cite{schmitt2019sen12ms}         & 180,662           & 17  & -  & $256\times256$     & SC, Seg     & MS \\
							  & BigEarthNet-MM \cite{sumbul2021bigearthnet}  & 590,326           & 43  & -  & $20\times20 \sim120\times120$        & SC                                                                     & MS \\
							  & Berlin \cite{yao2022multimodal}          & 464,671           & 8  & -    & $797\times220$     & IC                                                                     & HS \\
							  & SSL4EO-S12 \cite{wang2023ssl4eo}      & $\sim$3 million  & -  & -   & $264\times264$     & SC, Seg, CD & MS \\
							  & MMEarth \cite{nedungadi2024mmearth}         & $\sim$1.2 million & - & -   & $128\times128$     & IC, Seg     & Optical+MS \\ \hline
		\multirow{3}{*}{Hyperspectral}
							  & NeonTree \cite{weinstein2021benchmark}   & 457       & 2  & 5   & $400\times400$     & HTD    & Optical \\
							  & Airport \cite{zhou2010stable}            & 1       & 2  & 225   & $400\times400$     & HTD    & - \\
							  & Hermiston                                & 1       & 5  & 242   & $306\times241$     & CD    & - \\
							  & River \cite{wang2018getnet}              & 2       & -  & 198   & $463\times241$     & CD    & - \\
							  & Houston \cite{debes2014hyperspectral}    & 1       & 15 & 144   & $1905\times349$    & IC, HSR                                                                    & - \\
							  & Pavia \cite{billor2000bacon}             & 2       & 9  & 102, 103   & $1096\times1096$, $610\times340$     &  IC                                                                     & - \\ \hline
		\multirow{2}{*}{LiDAR} 
							  & MUUFL \cite{du2017technical}     & 53,687            & 11  &-  & $325\times220$     & IC                                                                     & MS \\
							  & DFC2018 \cite{le20182018}        & -                & 20   &-  & $8344\times2404$                  & IC                                                      & Optical+MS+HS \\ \hline		   
		\multirow{7}{*}{Vision-Language}
							  &UCM-Caption \cite{qu2016deep}   & 2,100/10,500         & 21    & -                        & $256\times256$   & ICa                      &- \\
							  &RS-CD \cite{lu2018exploring}    & 10,921/54,605        & 30    & -                        & $224\times224$   & ICa, ITR &- \\
							  &RS-TMD \cite{yuan2021exploring} & 4,743/23,715         & 32    & -                        & $224\times224$   & ICa, ITR &- \\
							  &DIOR-RSVG \cite{zhan2023rsvg}   & 17,402/38,320        & 20    & -                        & $800\times800$   & VG                       &- \\
							  &RSVG \cite{sun2022visual}       & 4,239/7,993          & 10    & -                        & $1024\times1024$ & VG                       &- \\
							  &RSVQA-HR \cite{lobry2020rsvqa}  & 10,659/955,664       & 89    & -                        & $512\times512$   & VQA               &-                     \\
							  &RSVQA-LR \cite{lobry2020rsvqa}  & 772/77,232           & 89    & -                        & $256\times256$   & VQA               &-                     \\ \hline
	\end{tabular}
\end{table*}

\subsubsection{SAR and Interferometric SAR (InSAR)}
This technology forms images by emitting microwaves and receiving echoes, offering all-weather, all-season imaging with strong penetration and sensitivity to surface roughness. Their texture, scattering, and structure differ from that of visible images, allowing reliable ground observation under clouds or at night. Thus, SAR is widely used in flood monitoring \cite{hamidi2023fast}, urban change detection \cite{manzoni2021joint}, and environmental monitoring \cite{asiyabi2023synthetic}.
Typical datasets are listed in Table~\ref{tab3}. Some, such as MMEarth \cite{nedungadi2024mmearth}, contain RGB, MS, and SAR for multimodal pretraining, spatially and temporally aligned via MAE.

\subsubsection{Hyperspectral (HS)}
This technology collects hundreds of contiguous narrow bands, yielding dense spectral signatures for each pixel. Such fine-grained information enables precise material identification via subtle spectral differences, supporting pixel-level discrimination for fine-grained land cover classification, vegetation analysis, and mineral exploration.
Common datasets for HS-based GFMs are listed in Table~\ref{tab3}. Each band corresponds to one spectral interval or data channel. Some methods unify channel numbers within batches or employ channel-embedding layers for dimensional alignment.
Because HS-based GFMs require diverse sensors and large training data, many rely on curated datasets for pretraining \cite{velazquez2025earthview,li2025hyperfree,wang2025hypersigma}, which are not elaborated here.

\subsubsection{LiDAR}
This technology measures distances via laser pulses, generating precise 3D point clouds or elevation rasters. Dataset statistics are summarized in Table~\ref{tab3}. For instance, DFC2018 \cite{le20182018} combines 48-band HS, 3-band LiDAR, and high-resolution RGB with GSDs of 1 m, 0.5 m, and 0.05 m, respectively. LiDAR complements optical sensors by providing geometric and elevation cues, enhancing spatial and structural understanding in RS applications.
Despite its spatial richness, LiDAR remains underutilized in GFMs due to structural mismatch and integration challenges. Nonetheless, it holds promise as a geometry-aware modality. He \textit{et al.} \cite{he2024foundation} jointly support HS, SAR, and LiDAR to mitigate cross-modal feature learning difficulties from heterogeneous pretraining sources.

\subsubsection{Vision-Language}
The development of VL-GFMs is driven by the domain-specific nature of RS data, high resolution, hierarchical tasks \cite{chen2025dynamicvis}, and non-standard visual semantics \cite{li2023rs}, which limit direct transfer from conventional vision-language models (VLMs). RS textual descriptions often include geographic knowledge and spatiotemporal metadata, demanding datasets that link image features with structured, domain-relevant semantics.
Such datasets aim to learn fine-grained correspondences between complex image features and specialized textual semantics. For open-vocabulary reasoning, they include multiple descriptive sentences covering spatial relationships, morphology, and context, offering richer supervision than categorical labels with higher annotation costs.
Given the large-scale data requirements of GFMs, text is often first generated by pretrained VLMs and refined by experts. For example, MMM-RS \cite{wang2024mmm} employs a pretrained VLM to produce textual prompts followed by manual correction, yielding high-quality image-text pairs with semantically rich supervision.

\subsection{Tasks and Performance}
\subsubsection{Tasks of Multimodal Geospatial Foundation Model}
We categorize multimodal GFM tasks into five visual and four image-text types to review multimodal GFMs.

\paragraph{Vision Tasks}
The tasks for GFM involve three levels: object-level, pixel-level, and scene-level.

\textbf{Object-level Tasks}:
Object detection (OD) locates and classifies ground objects and includes oriented and horizontal detection \cite{wang2025oriented}. The model requires suppression of complex backgrounds and robustness to object rotation.

\textbf{Pixel-level Tasks}:
Change detection (CD) identifies temporal differences in multi-temporal RS data \cite{dong2024changeclip} and distinguishes genuine changes from illumination, seasonal, or sensor variations.
Semantic segmentation (Seg) performs pixel-level classification for applications such as land cover mapping and urban planning \cite{ramos2024multispectral}. It demands determining multi-scale context and spectral similarity between different classes.

\textbf{Scene-level Tasks}:
Scene classification (SC) assigns a single semantic label to an entire image based on global patterns \cite{lv2022scvit}. The task has large intra-class appearance variations across geographies and scales.
Image classification (IC) assigns category labels to pixels or regions for land cover or land use mapping \cite{chen2024rsmamba}, requiring extraction of both low- and high-level features under diverse spatiotemporal conditions.

\paragraph{Vision-Language Tasks}
Image captioning (ICa) involves generating fluent and credible sentences that describe the given image's visual content \cite{wang2022end}. This task necessitates ensuring semantic alignment between image and text features.
Image-text retrieval (ITR) involves two sub-tasks: retrieving images based on natural language queries (text-to-image, T2I) and retrieving text descriptions for a given image (image-to-text, I2T). The structured nature of RS data and the unstructured language increase semantic alignment complexity \cite{yuan2023parameter}.
Visual grounding (VG) localizes specific image regions or objects based on natural language expressions \cite{zhan2023rsvg}, which is challenging in dense RS scenes that interpret precise directional, topological, and comparative expressions.
Visual question answering (VQA) requires comprehensive multimodal understanding, including dynamic reasoning, domain-specific semantic interpretation, and substantial RS domain knowledge \cite{rahnemoonfar2021floodnet, mi2024knowledge}. However, it remains difficult due to limited spatial context and strong domain dependence.

\subsubsection{Performance of Multimodal Geospatial Foundation Models}
We evaluate multimodal GFMs along two primary dimensions, modality composition and task specificity, comparing performance across vision-only and vision-language tasks to identify capability strengths and emerging trends.
\paragraph{Optical RGB}
\begin{table*}[h]
	\centering
	\caption{Object Detection Performance on Optical RGB Datasets.}
	\label{tab4}
	\begin{tabular}{llllllll}
	\hline
	Method      & Backbone  & \multicolumn{1}{c}{DIOR} & DIOR-R & \multicolumn{1}{c}{DOTAV1} & \multicolumn{1}{c}{DOTAV2} & FAIR1M & xView \\ \hline
	SeCo \cite{manas2021seasonal}        & ResNet-50 & -                        & -      & -                             & -                             & -          & 17.20 \\
	CACo \cite{mall2023change}        & ResNet-50 & 66.91                    & 64.10  & -                             & -                             & 47.83      & 17.20 \\
	GFM \cite{mendieta2023towards}         & Swin-B    & 72.84                    & 67.67  & -                             & -                             & 49.69      & -     \\
	SatMAE \cite{cong2022satmae}      & ViT-L     & 70.89                    & 65.66  & -                             & -                             & 46.55      & -     \\
	Scale-MAE \cite{reed2023scale}   & ViT-L     & 73.81                    & 66.47  & -                             & -                             & 48.31      & -     \\
	SatLas \cite{bastani2023satlaspretrain}      & Swin-B    & 74.10                    & 67.59  & -                             & -                             & 46.19      & -     \\
	SkySense \cite{guo2024skysense}    & Swin-H/ViT-L    & \underline{78.73}                    & \textbf{74.27}  & -                             & -                             & \textbf{54.57}      & -     \\
	Falcon \cite{yao2025falcon}      & -         & 77.01                    & -      & -                             & \underline{61.90}                         & -          & \textbf{27.81} \\
	CMID \cite{muhtar2023cmid}        & Swin-B    & 75.11                    & 66.37  & 77.36                         & -                             & 50.58      & -     \\
	MTP \cite{wang2024mtp}         & Swin-H    & \textbf{81.10}                    & \underline{72.17}  & \underline{80.77}          & 58.96                         & \underline{50.93}      & \underline{19.40}  \\
	RVSA \cite{wang2023advancing}        & ViTAE-B   & 73.22                    & 71.05  & \textbf{81.01}                         & \textbf{76.31}                         & 47.04      & -     \\
	RingMo \cite{sun2023ringmo}      & Swin-B    & 75.90                    & -      & -                             & -                             & 46.21      & -     \\ \hline
	\end{tabular}
\end{table*}
\begin{table*}[h]
	\centering
	\caption{Change Detection Performance on Optical RGB and Multispectral Datasets.}
	\label{tab5}
	\begin{tabular}{llllllllllllll}
	\hline
	\multirow{2}{*}{Methods}                     & \multirow{2}{*}{Backbone} & \multicolumn{4}{c}{LEVIR-CD (Optical RGB)}                                    & \multicolumn{4}{c}{WHU-CD (Optical RGB)}                                                   & \multicolumn{4}{c}{OSCD (Multispectral)}                                            \\ \cline{3-14} 
	                                             &                           & P                & R                & $F_{1}$    & mIoU        & P              & R              & $F_{1}$                    & mIoU         & P             & R                    & $F_{1}$             & mIoU  \\ \hline
	GFM \cite{mendieta2023towards}               & Swin-B                    & -                & -                & -           & -           & -              & -              & -                           & -            & \underline{58.07}         & \textbf{61.67}    & 59.82    & -     \\
	SpectralGPT \cite{hong2024spectralgpt}       & ViT-B                     & -                & -                & -           & -           & -              & -              & -                           & -            & 52.39 & \underline{57.20}                & 54.29                & -     \\
	Skysense \cite{guo2024skysense}              & Swin-H/ViT-L                    & -                & -                & 92.58       & -           & -              & -              & -                           & -            & -             & -                    & 60.06                & -     \\
	MTP-IMP \cite{wang2024mtp}                   & InternImage-XL            & -                & -                & 92.54       & -     & -              & -              & 95.59           & -            & -             & -                    & 55.61                & -     \\
	RSP \cite{wang2022empirical}            & IMP-ViTAEv2-S             & -                & -                & 91.26       & -           & -              & -              & -                           & -            & -             & -                    & -                    & -     \\
	RVSA \cite{wang2023advancing}                & ViT-B\&RVSA                & -                & -                & 90.86       & -           & -              & -              & -                           & -            & -             & -                    & 50.28                & -     \\
	SatMAE \cite{cong2022satmae}                 & ViT-L                     & -                & -                & 87.65       & -           & -              & -              & -                           & -            & 48.19         & 42.24                & 45.02                & -     \\
	RingMo-Aerial \cite{diao2025ringmo}          & RingMo-Aerial             & -                & -                & 92.71       & 87.10 & -         & -              & -                           & -            & -             & -                    & -                    & -     \\ 
	Meta-CD \cite{gao2025combining}              & CSPDarknet                & -                & -                & \textbf{95.53} & \underline{91.76} & -         & -              & \underline{97.52}           & \textbf{95.46} & -           & -                    & -                    & -     \\
	CSTSUNet \cite{wu2023cstsunet}               & ResNet18\&Swin             & 91.99            & 89.41            & 90.68       & 82.96       & -              & -              & -                           & -            & -             & -                    & -                    & -     \\
	ChangeCLIP \cite{dong2024changeclip}         & ResNet-50                 & 93.40            & \underline{90.67} & 92.01 & \textbf{92.18} & 95.63 & \underline{94.02} & 94.82  & 94.83 & - & -           & -                    & -     \\
	SAM-CD \cite{ding2024adapting}               & FastSAM-s                 & \textbf{95.87} & \textbf{95.14} & \underline{95.50} & 91.68 & \textbf{97.97} & \textbf{97.20} & \textbf{97.58} & \underline{95.36} & - & -           & \textbf{64.61}        & \textbf{73.47} \\
	DynamicVis \cite{chen2025dynamicvis}         & DynamicVis-L              & \underline{93.97}            & 90.48            & 92.32       & 85.31       & \underline{96.78} & 92.50 & 94.79 & 89.95 & \textbf{79.41} & 48.36 & \underline{60.25}       & \underline{43.16} \\ \hline
	\end{tabular}
\end{table*}

As the foundational modality derived from natural images, optical RGB remains the primary benchmark for object detection in GFMs. Table~\ref{tab4} compares representative methods using mean Average Precision (mAP), which averages precision scores across categories to reflect both classification and localization accuracy. The `-' symbol denotes unreported metrics, reflecting the models' design for broad task compatibility rather than specialized optimization.
Early approaches like MTP \cite{wang2024mtp} utilized SAM-generated pseudo-labels with multi-layer feature supervision to adapt decoders across semantic levels, achieving expected performance but constrained efficiency. This limitation has motivated the field's shift toward few-shot learning paradigms.

\paragraph{Optical RGB + Multispectral}
The combination of RGB and MS data (RGB-MS) is a common form in multimodal GFMs. Their high consistency in spatial structure and partial spectral overlap facilitate compatibility and transferability. The challenge lies not in modality heterogeneity but in adapting to variations in the spectral bands. Recent models address this obstacle while preserving spatial alignment. For example, SatMAE \cite{cong2022satmae} achieves RGB-MS compatibility by encoding MS into band groups with distinct spectral embeddings and shared temporal embeddings across patches.
In downstream tasks, RGB-MS models show promising results in change detection, semantic segmentation, and scene classification, as shown in Tables~\ref{tab5}-\ref{tab7}.

In CD, performance is evaluated comprehensively using precision (P), recall (R), $F_{1}$ score, and mean intersection over union (mIoU), which respectively reflect the model's ability to avoid false alarms, detect true changes, balance P-R trade-offs, and ensure spatial consistency. SAM-CD \cite{ding2024adapting} excels in such dynamic tasks, inheriting efficient encoding from FastSAM and introducing a residual convolution block in the decoder to construct change embeddings. This design extracts differential features across multi-temporal images and embeds discriminative semantics, facilitating accurate identification of surface changes.
In Seg, mIoU and mean $F_{1}$-score ($\mathrm{mF}_{1}$) are used for evaluation, accounting for varying class definitions across datasets. $\mathrm{mF}_{1}$ is critical for high-precision tasks like building or road segmentation, where both false positives and negatives must be minimized.
In SC, overall accuracy (OA) measures the proportion of correctly classified samples. OA 50\%/80\% denotes OA under a training ratio (TR) of 50\% or 80\%. Similarly, mAP 10\%/100\% indicates mean average precision at TR=10\% or 100\%. A smaller TR reflects few-shot learning performance. Top-1 accuracy (Acc) is the proportion of top-predicted classes matching the ground truth, while Top-5 Acc checks if the correct class is among the top five predictions.
For static tasks like segmentation and classification, the focus shifts to recognizing content or delineating regions. SkySense \cite{guo2024skysense} employs prototype learning by embedding spatiotemporal features into geographic coordinates, linking representative regions with semantic features. Its competitive performance also stems from a composite loss across image-, object-, and pixel-level semantics, enabling more precise and robust representation learning across diverse static tasks.

\paragraph{Optical RGB + SAR}
The integration of RGB and SAR data, referred to as RGB-SAR, constitutes a more challenging form of multimodal GFMs. Due to distinct imaging mechanisms, they differ substantially in spatial structures, texture features, and data scales. The main challenge lies in aligning heterogeneous physical properties and reconciling geometric-radiometric inconsistencies. As an early multimodal GFMs, CSPT \cite{zhang2022consecutive} adapts pre-trained ImageNet to the RS domain through task-aware training and task-agnostic MIM, leveraging task-relevant unlabeled data for continual pre-training to narrow the domain gap. Recent works further alleviate these inconsistencies via cross-modality feature alignment \cite{zhang2025skysense} and spatially contrastive learning \cite{fuller2023croma}.

Table~\ref{tab8} compares RGB-SAR performance across NWPU VHR-10 (object detection), SSDD (ship detection), and EuroSAT/EuroSAT-SAR (LULC classification). Among them, DynamicVis and FlexiMo yield competitive results. DynamicVis \cite{chen2025dynamicvis} employs meta-embeddings and multiple instance learning to disentangle heterogeneous distributions across million-scale datasets, distilling shared semantics within a dynamic region-aware architecture for efficient hierarchical representation. FlexiMo \cite{li2025fleximo} introduces a spatial-resolution-aware module for dynamic input calibration, preserving multi-scale fidelity, and a lightweight wavelength-guided channel adaptation mechanism that enhances multimodal consistency.
\begin{table*}[h]
	\setlength{\tabcolsep}{3pt}
	\centering
	\caption{Semantic Segmentation Performance on Optical RGB and Multispectral Datasets.}
	\label{tab6}
	\begin{tabular}{llccccccc}
	\hline
	\multirow{2}{*}{Methods} & \multirow{2}{*}{Backbone} & \multicolumn{4}{c}{Optical RGB}                                                                                                                                                                                                                                                                       & \multicolumn{3}{c}{Multispectral}                                                                  \\ \cline{3-9} 
	&                           & \begin{tabular}[c]{@{}c@{}}iSAID\\ mIoU\end{tabular} & \begin{tabular}[c]{@{}c@{}}Potsdam\\ $\mathrm{mF}_{1}$\end{tabular} & \begin{tabular}[c]{@{}c@{}}LoveDA\\ mIoU\end{tabular} & \begin{tabular}[c]{@{}c@{}}Vaihingen\\ mIoU\end{tabular} & \begin{tabular}[c]{@{}c@{}}Dyna.-Pla\\ mIoU (val/test)\end{tabular} & \begin{tabular}[c]{@{}c@{}}Dyna.-S2\\ mIoU (val/test)\end{tabular} & \begin{tabular}[c]{@{}c@{}}GID-15\\ mIoU\end{tabular} \\ \hline
	RingMoE \cite{bi2025ringmoe}                  & RingMoE(6.5B)             & \underline{69.70} & \underline{93.54}   & -      & -         & -                   & -/\textbf{47.60}            & -      \\
	SeCo \cite{manas2021seasonal}                     & ResNet-50                 & 57.20 & 89.03   & 43.63  & 68.90     & -                   & 29.40/39.80        & -      \\
	CACo \cite{mall2023change}                     & ResNet-50                 & 64.32 & 91.35   & 48.89  & -         & 35.40/42.70         & 30.20/42.50        & -      \\
	GFM \cite{mendieta2023towards}                      & Swin-B                    & 66.62 & 91.85   & -      & \underline{75.30}     & 36.70/\underline{45.60}         & -                  & -      \\
	CMID \cite{muhtar2023cmid}                     & Swin-B                    & 66.21 & 91.86   & -      & -         & 36.40/43.50         & -                  & -      \\
	RVSA \cite{wang2023advancing}                     & ViTAE-B                   & 64.49 & -       & \textbf{52.44}  & 71.60     & 34.30/44.40         & -                  & \textbf{92.60}  \\
	SatMAE \cite{cong2022satmae}                   & ViT-L                     & 62.97 & 90.63   & -      & 69.30     & 32.80/39.90         & 30.10/38.70        & 88.10  \\
	SatLas \cite{bastani2023satlaspretrain}                   & Swin-B                    & 68.71 & 91.28   & -      & -         & \underline{37.40}/40.70         & \underline{31.90}/43.50        & -      \\
	SkySense \cite{guo2024skysense}                 & Swin-H/ViT-L                    & \textbf{70.91} & \textbf{93.99}   & -      & -         & \textbf{39.70}/\textbf{46.50}         & \textbf{33.10}/\underline{46.20}        & -      \\
	Scale-MAE \cite{reed2023scale}                & ViT-L                     & 65.77 & 91.54   & -      & 72.30     & 34.00/41.70         & -                  & \underline{90.70}  \\
	RingMo \cite{sun2023ringmo}                   & Swin-B                    & 67.20 & 91.27   & -      & -         & -                   & -                  & -      \\
	Ringmo-SAM \cite{yan2023ringmo}               & ViT-L                     & 57.99 & -       & -      & 72.44     & -                   & -                  & -      \\
	FTUNetformer+SAM \cite{wang2022unetformer}         & Swin-B                    & -     & -       & \underline{50.68}  & \textbf{84.01}     & -                   & -                  & -      \\ \hline
	\end{tabular}
\end{table*}
\begin{table*}[h]
	\centering
	\caption{Scene Classification Performance on Optical RGB and Multispectral Datasets.}
	\label{tab7}
	\begin{tabular}{llcccccc}
	\hline
	\multirow{2}{*}{Methods} & \multirow{2}{*}{Backbone} & \multicolumn{3}{c}{Optical RGB}                                                                                                                                                                                                                 & \multicolumn{3}{c}{Multispectral}                                                                                                                                                                      \\ \cline{3-8} 
								 &                           & \multicolumn{1}{c}{\begin{tabular}[c]{@{}c@{}}UCM\\ OA 50/80\%\end{tabular}} & \multicolumn{1}{c}{\begin{tabular}[c]{@{}c@{}}AID\\ OA 20/40\%\end{tabular}} & \multicolumn{1}{c}{\begin{tabular}[c]{@{}c@{}}RESISC45\\ OA 10/20\%\end{tabular}} & \multicolumn{1}{c}{\begin{tabular}[c]{@{}c@{}}EuroSAT\\ OA 100\%\end{tabular}} & \multicolumn{1}{c}{\begin{tabular}[c]{@{}c@{}}BEN\\ mAP 10/100\%\end{tabular}} & \begin{tabular}[c]{@{}c@{}}fMoW-s2\\ Top1/5 Acc\end{tabular} \\ \hline
	RoMA \cite{wang2025roma}                    & Mamba-B        & 59.45/-           & -/87.36            & -                    & -      & -            & -                        \\
	SatMAE \cite{cong2022satmae}                & ViT-L          & -                 & 95.02/96.94        & 91.72/94.10          & 95.74  & 86.18/\underline{89.50}  & \underline{63.84}/-                         \\
	Scale-MAE \cite{reed2023scale}              & ViT-L          & -                 & 96.44/97.58        & 92.63/95.04          & \underline{98.59}  & -            & - \\
	GFM \cite{mendieta2023towards}              & Swin-B         & \underline{99.00}/-           & 95.47/97.09        & 92.73/94.64          & -      & \underline{86.30}/-      & - \\
	RVSA \cite{wang2023advancing}               & ViT-B          & \textbf{99.56}/-           & 97.03/\underline{98.48}        & 93.93/\underline{95.69}          & -      & -            & - \\
	UPetu \cite{dong2024upetu}                  & ConvNeXt       & 98.52/99.05       & 96.29/97.06        & 92.13/93.79          & -      & -            & - \\
	SeCo \cite{manas2021seasonal}               & ResNet-50      & -/97.86           & 93.47/95.99        & 89.64/92.91          & 97.34  & 82.62/87.81  & 51.65/77.40 \\
	RSP \cite{wang2022empirical}           & IMP-ViTAEv2-S  & -/\textbf{99.90}           & 96.91/98.22        & 94.41/95.60          & -      & -            & - \\
	RingMo \cite{sun2023ringmo}                 & Swin-B         & -/\underline{99.84}           & 96.90/98.34        & 94.25/95.67          & -      & -            & - \\
	RingMoE \cite{bi2025ringmoe}                & RingMoE(6.5B)  & -                 & \textbf{98.19}/-            & \textbf{95.90}/-              & -      & -            & - \\
	RingMo-Aerial \cite{diao2025ringmo}         & RingMo-Aerial  & -                 & 95.81/96.46        & 92.28/95.65          & -      & -            & - \\
	SatLas \cite{bastani2023satlaspretrain}     & Swin-B         & -                 & 94.96/97.38        & 92.16/94.70          & \textbf{98.97}  & 82.80/88.37  & 57.95/\underline{79.00}\\
	SkySense \cite{guo2024skysense}             & Swin-H/ViT-L         & -                 & \underline{97.68}/\textbf{98.60}        & \underline{94.85}/\textbf{96.32}          & -      & \textbf{88.67}/\textbf{92.09}  & \textbf{64.38}/\textbf{87.27}\\ \hline
	\end{tabular}
\end{table*}
\paragraph{Optical RGB + Hyperspectral}
Optical RGB contains only three spectral bands, whereas HS data encompasses tens to hundreds of contiguous narrow bands, leading to modality asymmetry, spectral redundancy, and heightened noise sensitivity. These issues, along with mismatched spatial and spectral resolutions, hinder fine-grained representation, particularly for pixel-level tasks. Unlike SAR or LiDAR, the fusion challenge here lies primarily in the spectral domain.
Despite this, RGB-HS-based GFMs have gained traction. LDGNet \cite{zhang2023language} learns generalizable features via cross-domain priors and language-grounded semantics, enabling zero-shot hyperspectral classification. DOFA \cite{xiong2024neural} draws on neural plasticity, introducing a dynamic hypernetwork that integrates multi-dataset inputs into a unified feature space through adaptive embedding across channel dimensions.

As summarized in Table~\ref{tab8}, diverse datasets support these tasks: NeonTree provides pixel-level forest annotations; Hermiston\footnote{Heriston: \url{https://earthexplorer.usgs.gov/}} serves for change detection, Houston for image classification, and MUUFL offers co-registered LiDAR-HS data for land cover analysis. For Houston, OA, Average Accuracy (AA), and the Kappa Coefficient (KC) are standard metrics—OA gauges overall correctness but is imbalance-sensitive; KC accounts for chance agreement; and AA balances per-class accuracy to mitigate skew.
Recent advances further extend hyperspectral interpretation. HyperSIGMA \cite{wang2025hypersigma} employs sparse sampling attention and a spectral enhancement module, supported by large-scale pre-training for effective cross-modal transfer. SatDiFuser \cite{jia2025can} leverages multi-stage diffusion representations and introduces global, local, and mixture-of-experts fusion to capture multi-grained semantics, achieving good results on NeonTree and EuroSAT.
\begin{table*}[!h]
	\setlength{\tabcolsep}{3pt}
	\centering
	\caption{Performance Evaluation on Optical RGB, SAR, Hyperspectral, and LiDAR in Multimodal Geospatial Foundation Models. \\The `$\text{Acc}^{1}$' Represents the Top1-Acc.}
	\label{tab8}
	\begin{tabular}{llcccccccc}
	\hline
	\multirow{2}{*}{Methods}                                      & \multirow{2}{*}{Backbone} & \multicolumn{2}{c}{Optical RGB}                                                                                               & \multicolumn{2}{c}{SAR}                                                                                         & \multicolumn{3}{c}{Hyperspectral}                                                                                                                                                         & LiDAR                                \\ \cline{3-10} 
																	&                           & \begin{tabular}[c]{@{}c@{}}EuroSAT\\ OA/AA/$\text{Acc}^{1}$\end{tabular} & \begin{tabular}[c]{@{}c@{}}NWPUVHR\\ mAP50\end{tabular} & \begin{tabular}[c]{@{}c@{}}SSDD\\ mAP50\end{tabular} & \begin{tabular}[c]{@{}c@{}}EuroSAT-SAR\\ OA\end{tabular} & \begin{tabular}[c]{@{}c@{}}NeonTree\\ mIoU\end{tabular} & \begin{tabular}[c]{@{}c@{}}Hermiston\\ IoU/OA/$F_{1}$\end{tabular} & \begin{tabular}[c]{@{}c@{}}Houston\\ OA/AA/KC\end{tabular} & \begin{tabular}[c]{@{}c@{}}MUUFL\\ OA/AA/KC\end{tabular} \\ \hline
	RSPrompter \cite{chen2024rsprompter}    & SAM \& ViTs               & -                        & \underline{91.70}            & \underline{95.60}         & -                  & -                 & -                                 & -                                                     & -                                                        \\
	CSPT \cite{zhang2022consecutive}        & ViT-B                     & -                        & 88.90                        & 91.80                     & -                  & -                 & -                                 & -                                                     & -                                                        \\
	DynamicVis \cite{chen2025dynamicvis}    & DynamicVis-L              & -                        & \textbf{91.90}               & \textbf{97.80}            & -                  & -                 & -                                 & -                                                     & -                                                        \\
	FlexiMo \cite{li2025fleximo}            & ViT-B                     & \textbf{99.44}/-/-       & -                            & -                         & \textbf{89.96}     & -                 & -                                 & -                                                     & -                                                        \\
	SeaMo \cite{li2025seamo}                & ViT-B                     & -/-/\underline{99.37}    & -                            & -                         & \underline{89.69}  & -                 & -                                 & -                                                     & -                                                        \\
	DOFA \cite{xiong2024neural}             & ViT-B                     & -/-/92.20                & -                            & -                         & 88.59              & 58.60             & -                                 & -                                                     & -                                                        \\
	CROMA \cite{fuller2023croma}            & ViT-B                     & -/-/\textbf{99.46}       & -                            & -                         & 88.42              & 56.30             & -                                 & -                                                     & -                                                        \\
	Scale-MAE \cite{reed2023scale}          & ViT-L                     & -/-/78.90                & -                            & -                         & -                  & 51.00             & -                                 & -                                                     & -                                                        \\
	GFM \cite{mendieta2023towards}          & Swin-B                    & -/-/89.20                & -                            & -                         & -                  & 51.00             & -                                 & -                                                     & -                                                        \\
	RemoteCLIP \cite{liu2024remoteclip}     & ViT-B                     & -/\underline{30.74}/-    & -                            & -                         & -                  & 56.20             & -                                 & -                                                     & -                                                        \\
	GeoLangBind \cite{xiong2025geolangbind} & ViT-L                     & -/\textbf{52.30}/-       & -                            & -                         & -                  & \underline{59.00} & -                                 & -                                                     & -                                                        \\
	SatDiFuser \cite{jia2025can}            & DiffusionSat              & -/-/97.70                & -                            & -                         & -                  & \textbf{63.80}    & -                                 & -                                                     & -                                                        \\
	LDGNet \cite{zhang2023language}         & ViT-B                     & -                        & -                            & -                         & -                  & -                 & -                                 & 80.34/-/65.80                                         & -                                                        \\
	HyperFree \cite{li2025hyperfree}        & ViT-B                     & -                        & -                            & -                         & -                  & -                 & \textbf{58.51}/-\underline{73.82} & -                                                     & -                                                        \\
	HyperSIGMA \cite{wang2025hypersigma}    & ViT-B                     & -                        & -                            & -                         & -                  & -                 & -\textbf{96.24}/\textbf{92.08}    & \underline{87.33}/\underline{88.74}/\underline{86.30} & -                                                        \\
	SSFTT \cite{sun2022spectral}            & CNN                       & -                        & -                            & -                         & -                  & -                 & -                                 & 87.04/87.38/85.98                                     & \underline{82.56}/\textbf{58.69}/\underline{76.57}                                  \\
	FMA \cite{he2023foundation}             & ViTs                      & -                        & -                            & -                         & -                  & -                 & -                                 & \textbf{90.35}/\textbf{90.11}/\textbf{89.57}          & \textbf{86.77}/\underline{57.56}/\textbf{82.16}                                  \\ \hline
	\end{tabular}
\end{table*}
\begin{table*}[!h]
	\centering
	\caption{Performance Comparison of Vision-Language Geospatial Foundation Models.}
	\label{tab9}
	\begin{tabular}{lccccccc}
		\hline
		\multirow{2}{*}{Methods} & \multicolumn{7}{c}{Image Captioning on the UCM-Caption}                                                                                                                                          \\ \cline{2-8} 
								 & BLEU-1                    & BLEU-2                    & BLEU-3                    & BLEU-4                    & METEOR                    & ROUGE\_L                     & CIDEr                      \\ \hline
		RSGPT \cite{hu2025rsgpt}                    & 86.12                     & 79.14                     & 72.31                     & 65.74                     & 42.21                     & 78.34    & 333.23 \\
		RS-CapRet \cite{silva2024large}                & 84.30                     & 77.90                     & 72.20                     & 67.00                     & \underline{47.20}                     & \underline{81.70}    & 354.80 \\
		SkyEyeGPT \cite{zhan2025skyeyegpt}                & \underline{90.70}                     & \underline{85.70}                     & \textbf{81.60}                     & \textbf{78.40}                     & 46.20                     & 79.50    & 236.80 \\
		RS-MoE \cite{lin2025rs}                   & \textbf{94.81}                     & \textbf{87.09}                     & \underline{79.57}                     & \underline{72.34}                     & \textbf{66.97}                     & 62.74    & \textbf{396.46} \\
		BITA \cite{yang2024bootstrapping}                     & 87.27                     & 80.96                     & 75.51                     & 70.39                     & 46.52                     & \textbf{82.58}    & \underline{371.29} \\
		EarthGPT \cite{zhang2024earthgpt}                 & 87.10 & 78.70 & 71.60 & 65.50 & 44.50 & 78.20    & 192.60 \\ \hline
		\multirow{2}{*}{Methods} & \multicolumn{7}{c}{Text-to-Image Retrieval on the RS-CD}                                                                                                                                      \\ \cline{2-8} 
								 & I2T @R1                   & I2T @R5                   & I2T @R10                  & T2I @R1                   & T2I @R5                   & \multicolumn{1}{l}{T2I @R10} & \multicolumn{1}{l}{mean R} \\ \hline
		KCR \cite{mi2022knowledge}                      & 5.40                      & 22.44                     & 37.36                     & 5.95                      & 18.59                     & 29.58                        & 19.89                      \\
		RemoteCLIP \cite{liu2024remoteclip}               & \underline{18.39}                     & \underline{37.42}                     & \underline{51.05}                     & \underline{14.73}                     & \underline{39.93}                     & \underline{56.58}                        & \underline{36.35}                      \\
		RS-CapRet \cite{silva2024large}                & -                         & -                         & -                         & 10.25                     & 31.62                     & 48.53                        & 30.13                      \\
		SkyCLIP \cite{wang2024skyscript}                  & 8.97                      & 24.15                     & 37.97                     & 5.85                      & 20.53                     & 33.53                        & 21.84                      \\
		GeoLangBind \cite{xiong2025geolangbind}              & 8.42                      & 25.16                     & 37.05                     & 8.15                      & 24.90                     & 37.73                        & -                          \\
		GeoRSCLIP \cite{zhang2024rs5m}                      & \textbf{21.13}                      & \textbf{41.72}                     & \textbf{55.63}                     & \textbf{15.59}                      & \textbf{41.19}                     & \textbf{57.99}            & \textbf{38.87}                      \\ \hline
		\multirow{2}{*}{Methods} & \multicolumn{7}{c}{Visual Grounding on the DIOR-RSVG}                                                                                                                                         \\ \cline{2-8} 
								 & P@0.5                     & P@0.6                     & P@0.7                     & P@0.8                     & P@0.9                     & mIoU                         & cumIoU                     \\ \hline
		LHRS-Bot \cite{muhtar2024lhrs}                 & \textbf{88.10}                     & -                         & -                         & -                         & -                         & -                            & -                          \\
		Falcon \cite{yao2025falcon}                   & \underline{87.54}                     & -                         & -                         & -                         & -                         & -                            & -                          \\
		TransVG \cite{deng2021transvg}                  & 72.41                     & 67.38                     & 60.05                     & 49.10                     & 20.91                     & 58.57                        & 76.27                      \\
		RSVG \cite{zhan2023rsvg}                    & 76.78                     & \textbf{72.68}                     & \textbf{66.74}                     & \underline{56.42}                     & \underline{35.07}                     & \underline{68.04}                        & \underline{78.41}                      \\
		EarthGPT \cite{zhang2024earthgpt}                 & 76.65                     & \underline{71.93}                     & \underline{66.52}                     & \textbf{56.53}                     & \textbf{37.63}                     & \textbf{69.34}                        & \textbf{81.54}                      \\ \hline
		\multirow{3}{*}{Methods} & \multicolumn{7}{c}{Visual Question Answering}                                                                   \\ \cline{2-8} 
								 & \multicolumn{3}{c}{on the RSVQA-HR (test set2)} & \multicolumn{4}{c}{on the RSVQA-LR}                                \\
								 & Presence     & Comparison    & Avg. Acc    & Presence & Comparison & \multicolumn{1}{l}{Rural/Urban} & Avg. Acc \\ \hline
		RSVQA \cite{lobry2020rsvqa}                    & 86.26        & 85.94         & 86.10       & 87.46    & 81.50      & \underline{90.00}                           & 86.32    \\
		RSGPT \cite{hu2025rsgpt}                    & \underline{89.87}        & \underline{89.68}         & \underline{89.78}       & \textbf{91.17}    & \textbf{91.70}      & \textbf{94.00}                           & \textbf{92.29}    \\
		SkyEyeGPT \cite{zhan2025skyeyegpt}                & 83.50        & 80.28         & 81.89       & 88.93    & 88.63      & 75.00                           & 84.19    \\
		GeoChat \cite{kuckreja2024geochat}                  & 58.45        & 83.19         & 70.82       & \underline{91.09}    & \underline{90.33}      & \textbf{94.00}                           & \underline{91.81}    \\
		EarthGPT \cite{zhang2024earthgpt}                 & 62.77        & 79.53         & 72.06       & -        & -          & -                               & -        \\
		LHRS-Bot \cite{muhtar2024lhrs}                 & \textbf{92.57}        & \textbf{92.53}         & \textbf{92.55}       & 88.51    & 90.00      & 89.07                           & 89.19    \\ \hline
	\end{tabular}
\end{table*}

\paragraph{Hyperspectral + LiDAR}
Unlike optical RGB that captures appearance and texture, LiDAR focuses on height and spatial contours, leading to significant discrepancies in data structure and distribution. These cross-modal differences introduce challenges such as semantic misalignment, projection errors, and inconsistent feature processing, compounded by LiDAR's inherent sparsity, irregularity, and lack of standardized resolution.
Nevertheless, growing demand for 3D scene understanding—in urban modeling, terrain analysis, and building recognition—has spurred preliminary integration of LiDAR into GFMs. For instance, msGFM \cite{han2024bridging} incorporates DSM data with a decoupled architecture, using separate patch embeddings and decoders per modality to enhance semantic segmentation on the Vaihingen dataset.
Compared to extensively supported modalities like HS and SAR, few GFMs demonstrate effective RGB-LiDAR fusion. As shown in Table~\ref{tab8}, FMA \cite{he2023foundation} proposes a tuning-free multimodal adaptation framework that extracts spatial-channel features and models explicit functional dependencies between modalities.
As a key source of structural information, LiDAR offers considerable geometric value and adaptation potential. However, given its heterogeneity with RGB, further research remains urgently needed in feature alignment, modality projection, and 3D reconstruction.

\paragraph{Vision-Language}
We evaluate model performance and underlying mechanisms across cross-modal understanding and generation tasks. Tables~\ref{tab3} and~\ref{tab9} provide a comprehensive analysis of four representative VL-GFM tasks, covering sentence-level generation, region-level localization, and question-level reasoning.
As shown in Table~\ref{tab9}, the ICa task involves BLEU-1 to BLEU-4, METEOR, ROUGE-L, and CIDEr: BLEU-$n$ evaluates $n$-gram fluency, METEOR captures semantic similarity using synonymy and morphology, and ROUGE-L measures structural overlap via the longest common subsequence. For the ITR, R@1 scores reflect the accuracy of retrieving text from an image (I2T) or an image from text (T2I). In the VG, cumulative IoU (cumIoU) aggregates intersection-over-union across all predictions as a global accuracy measure, while Pr@0.5 evaluates localization precision at IoU $\geq$ 0.5. The VQA performance is assessed using Average Accuracy (Avg. Acc) and per-type accuracy for questions such as Presence (yes/no), Comparison, and Rural or Urban classification.

Overall, no single universal model dominates; instead, architectures and objectives yield complementary strengths.
ICa requires object detection, relation inference, and fluent description. RS-MoE \cite{lin2025rs} introduces an instruction router that dynamically generates task-specific prompts to guide multiple lightweight LLM experts, enabling specialized parallel processing of diverse VL subtasks.
ITR relies on aligning text and images in a shared embedding space. GeoRSCLIP \cite{zhang2024rs5m}, fine-tuned on the linguistically diverse RS5M dataset, achieves strong alignment and retrieval performance.
VG demands fine-grained spatial understanding and language reasoning. EarthGPT \cite{zhang2024earthgpt} integrates a visually enhanced perception module for multi-scale details with a cross-modal comprehension strategy to align linguistic references with visual regions.
VQA combines visual understanding with semantic reasoning. RSGPT \cite{hu2025rsgpt}, fine-tuned on the RSICap dataset with optimized Q-Former and linear layers, achieves accurate multimodal alignment and inference on scenes, attributes, and spatial relations.
\begin{figure*}[t]
	\centering
	\includegraphics[width=\textwidth]{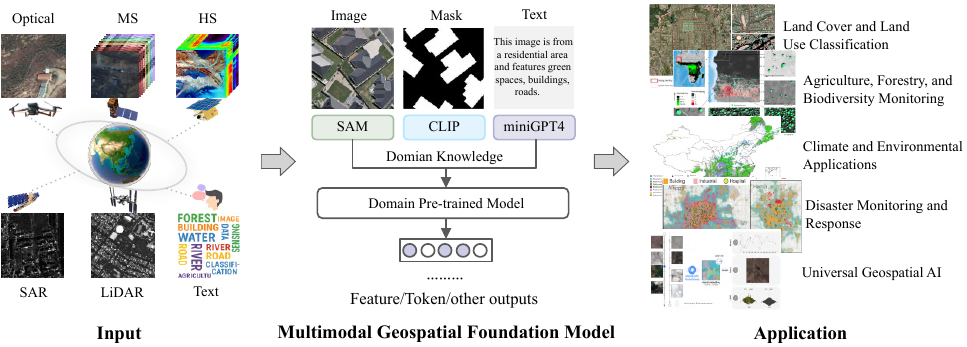}
	\caption{Applications of multimodal geospatial foudation models. Land cover and land use classification: urban planning \cite{zhang2025cmab}; agriculture, forestry, and biodiversity monitoring: global tree monitoring \cite{brandt2020unexpectedly}; climate and environmental applications: carbon emission estimation \cite{liao2024growing}; disaster monitoring and response: post-war damage analysis \cite{hou2024war}; universal geospatial AI: virtual satellite AI model from Google \cite{brown2025alphaearth}.}
	\label{figure6}
\end{figure*}
Broadly speaking, VL-GFMs represent a key trend in intelligent remote sensing interpretation. VL-GFMs show us natural language as a semantic bridge for human-machine interaction. VL-GFMs not only enable a cognitive shift from low-level pixel features to high-level semantic understanding, but also highlight cross-modal representations for generalization across diverse mainstream tasks. In particular, their multi-task collaborative frameworks, such as the complementarity between image-text retrieval and visual grounding, demonstrate strong interpretability and adaptability in complex scenarios. This paradigm shift allows RS systems to achieve both professional-level accuracy and intuitive natural interactions.

\section{Applications of Multimodal Geospatial Foundation Models}
The emergence of multimodal GFM represents a paradigm shift from isolated, modality-specific analysis to an integrated understanding of Earth observations. As shown in Fig.~\ref{figure6}, these models, pretrained on five core RS modalities or vision-language data, encode not only low-level physical characteristics unique to each sensor but, more importantly, high-level semantic concepts and cross-modal correlations. This enables them to capture complex intermodal interactions. The resulting unified representation forms a powerful, transferable feature space that, with minimal fine-tuning, achieves strong robustness and adaptability across diverse downstream applications. The potential of multimodal GFMs is illustrated in the following five domains.
\subsection{Land Cover and Land Use Classification}
As a key application in static RS analysis, land cover and land use (LCLU) classification is essential for environmental understanding. When combined with dynamic RS data, accurate LCLU maps underpin agricultural monitoring, infrastructure planning, and sustainable development. Common categories include vegetation, water, impervious surfaces, and bare land. HS data provide additional spectral cues, sensitive to surface physical and chemical properties, helping to distinguish objects with similar spectra, such as crops, bare soil, and asphalt.
CMAB \cite{zhang2025cmab} integrates RS and street-view data for nationwide, multi-attribute building extraction, extending conventional 2D mapping to 3D functional characterization for urban planning. Hanan \textit{et al.} \cite{hanan2020satellites} explored using high-resolution satellite imagery and foundation models for global-scale tree census initiatives. A persistent challenge for multimodal GFMs in LCLU tasks is large intra-class scale variation. ScaleMAE \cite{reed2023scale} mitigates this issue by area-aware positional encoding and Laplacian pyramid-based decomposition, enhancing multiscale object recognition.
\subsection{Agriculture, Forestry, and Biodiversity Monitoring}
Unlike broad LCLU classification, recent efforts emphasize fine-grained categories and temporal dynamics, revealing new scientific insights. Zhu \textit{et al.} \cite{brandt2020unexpectedly} used MS data to identify non-forest trees across the Sahara and Sahel, advancing understanding of land degradation and carbon sinks. These applications demand finer class definitions, higher resolution, and denser temporal sampling, often requiring domain-specific knowledge such as crop rotation patterns or spectral libraries of tree species. Optical signals indicate vegetation nutrient levels; near-infrared reflectance correlates with nitrogen, while red-band reflectance relates to phosphorus \cite{gholizadeh2019detecting,janga2023review}, and SAR complements them by capturing surface roughness and soil moisture.
For multimodal GFMs, such heterogeneity complicates the modeling of periodic and structural details. FoMo \cite{bountos2025fomo} addresses these challenges through dynamic multispectral masking and multi-scale gradient optimization, demonstrating broad applicability in forest monitoring.
\subsection{Climate and Environmental Applications}
Climate and environmental applications represent a highly interdisciplinary and widely influential domain in RS, encompassing multi-scale tasks ranging from local ecosystem monitoring to global climate change assessment. Critical for analyzing climate variability, detecting environmental degradation, and enabling global carbon accounting efforts, these applications exemplify the strategic value of Earth observation in addressing planetary-scale challenges.
Liao \textit{et al.} \cite{liao2024growing} employed optical and LiDAR remote sensing technologies to quantify China's woody biomass carbon sinks, providing a critical satellite-independent verification for the greenhouse gas inventory and a scientific basis for evaluating and guiding nature-based climate solutions.
FengWu-GHR \cite{han2024fengwu} developed a global weather forecasting system based on high-resolution pretrained models, achieving a spatial resolution of 0.09°, thus facilitating medium- and long-term meteorological modeling. Additionally, ClimateBERT \cite{webersinke2021climatebert} introduced a pretrained language model tailored for climate-related textual data, advancing cross-modal climate intelligence research.
\subsection{Disaster Monitoring and Response}
Disaster monitoring and response are vital for emergency management and public safety, supporting early warning, situational awareness, and post-disaster assessment across hazards like earthquakes, floods, and wildfires. These applications require high spatiotemporal responsiveness and strong generalization.
Recent efforts leverage multimodal data and rapid processing: the European Centre for Medium-Range Weather Forecasts (ECMWF) \cite{ECMWF2021} promotes integrated systems for hazard prediction; FloodNet \cite{rahnemoonfar2021floodnet} fuses unmanned aerial vehicle (UAV) imagery and multi-task annotations for flood assessment; FMARS \cite{arnaudo2024fmars} employs a GFM to integrate open-source disaster data for automatic damage dataset generation; and Hou \textit{et al.} \cite{hou2024war} propose temporal knowledge-guided detection for near-real-time building damage estimation. These advances mark a transition toward multi-task, multimodal, and time-critical disaster modeling.
\subsection{Universal Geospatial AI}
Following the success of general FMs such as ChatGPT and DeepSeek, AI democratization is accelerating their adoption across sectors, including education, healthcare, and public services. Building general-purpose GFMs is thus a key research direction. They are envisioned to serve not only RS and AI researchers but also practitioners such as conservationists, planners, and emergency agencies.
Open geospatial AI should support low- or zero-code interaction and seamless deployment through libraries like Raster Vision and TorchGeo \cite{stewart2022torchgeo}. New data efforts integrate OpenStreetMap tags with Google Earth Engine imagery for RS image-text annotations and VQA tasks \cite{sun2022visual}, while GPT-4V-based frameworks generate large-scale multimodal datasets \cite{yuan2024chatearthnet,zhang2024rs5m}. Commercial multimodal GFMs such as RingMo and Cangling Brain exhibit strong transferability across diverse tasks. AlphaEarth \cite{brown2025alphaearth} introduces an embedding field model to transform sparse annotations into generalizable geospatial representations, building a universal foundation for global-scale analysis.
Future work can develop domain-specific applications for non-technical users, promoting the broad adoption of geospatial AI in scientific discovery and decision-making.

\section{Challenges and Future Directions}
\subsection{Multimodal Domain Generalization and Imbalance Data}
RS data are inherently heterogeneous, encompassing multi-source, multimodal, and multi-temporal properties. Their spatial distribution varies widely across regions, leading to intra- and inter-domain shifts that limit model generalization. This issue intensifies under few-shot settings due to long-tailed category distributions, causing class imbalance during training. As the relative importance of modalities differs by task, current methods lack a unified framework to adaptively balance multimodal disparities. Moreover, the absence of quantitative metrics for data quality and distribution uniformity constrains optimization of data scheduling and training strategies.

Future research should develop hierarchical evaluation systems for RS datasets and incorporate statistical or learning-based measures of domain distance to guide data scheduling and improve strategies. Such approaches can enhance robustness under limited data. Addressing multimodal and cross-domain generalization also requires modeling task relevance and building collaborative adaptation mechanisms. A unified framework for multi-task modality adaptation could yield more robust and semantically consistent RS understanding.
\subsection{Model Interpretability, Trustworthiness, and Responsible AI}
RS data are increasingly applied in high-stakes domains such as environmental monitoring, climate change assessment, and disaster response, where decisions carry significant societal and ecological consequences. Such contexts demand not only transparency, traceability, and robustness but also the integration of Responsible AI (RAI) principles throughout the full model lifecycle \cite{ghamisi2025responsible}. However, most state-of-the-art GFMs are trained in an end-to-end fashion, often with limited interpretability, which constrains their ability to provide verifiable and trustworthy insights for domain experts in complex scenarios \cite{DoshiVelez2017, Rudin2019, Samek2021}.

Future work should move beyond accuracy metrics to integrate RAI principles into the design, training, and deployment of GFMs. Embedding geoscientific knowledge into large-scale architectures can enhance interpretability at semantic, structural, and physical levels \cite{Reichstein2019}. Promising directions include knowledge graph-guided attention, multiscale causal graph modeling, and other methods that enable hierarchical reasoning and semantic transparency \cite{Miller2019}. In parallel, standardized measures of trustworthiness are needed to quantify stability, fairness, and robustness across diverse geospatial conditions \cite{Zhang2023}. Adopting such practices will enable GFMs to evolve into responsible, explainable, and accountable systems that ensure credibility and ethical alignment from data collection to policy translation.
\subsection{Computational and Deployment Costs}
GFMs have to process large-scale, high-resolution imagery with massive parameters, leading to substantial computational demands. Beyond accuracy, deployment feasibility across aerial, terrestrial, and satellite platforms is critical, particularly regarding edge device compatibility and cost. HS data contain redundant spectral bands, and model knowledge transfer can introduce parameter redundancy, reducing efficiency and inflating deployment costs.

Lightweight multimodal GFM research should emphasize pruning, knowledge distillation, or mixed-precision computation to compress inference and improve efficiency. Edge computing and cloud-edge collaboration can enhance applicability in resource-limited settings, while cross-platform standardization will support sustainable deployment. Evaluation should also include energy and carbon metrics to align model scalability with environmental goals. Although concerns about energy usage persist, fine-tuning remains a more efficient pathway \cite{ghamisi2025geospatialFM}. For example, full fine-tuning of SpectralGPT emits 270 g of $\text{CO}_2$—equivalent to 0.7 miles of driving and under 1\% of the U.S. daily per-person average (39 kg/day) \cite{epa_vehicle_emissions}. Compared with training from scratch, fine-tuning leverages pretrained knowledge for efficient adaptation, making it both energy-saving and scalable.
\subsection{Ethical and Privacy Concerns}
High-resolution RS data face restrictions from commercial copyrights, national sovereignty, and privacy laws, severely limiting data sharing. Moreover, sensitive information such as military or personal activity patterns risks data leakage and ethical violations. These factors hinder model generalization and expose deployed AI systems to threats including poisoning, tampering, and unauthorized access. The absence of standardized data protocols further limits reproducibility and cross-institutional collaboration.
Developing open geospatial AI models is essential but needs to be grounded in strong privacy protection. Future efforts should improve standardization of public RS platforms and establish secure, compliant access mechanisms. Simultaneously, a regulatory framework for geospatial AI ethics should cover data provenance, output auditing, and risk assessment to ensure trustworthiness and control in open-access contexts.
\section{Conclusion}
The emergence of foundation models has reshaped remote sensing, shifting the field from isolated, task-specific pipelines toward unified, general-purpose frameworks capable of cross-modal reasoning. This review systematically traces the evolution of geospatial foundation models from a modality-centric perspective, highlighting how the intrinsic characteristics of each data type fundamentally shape model design. We synthesize core technologies in training paradigms and architectures that address key challenges like modality heterogeneity and the semantic gap, while large-scale benchmarks have been essential for rigorous evaluation. These models are already creating tangible societal impact, from precision agriculture to disaster response.
Nevertheless, the journey toward general-purpose geospatial AI continues. Critical challenges remain in domain generalization, model interpretability for high-stakes decisions, and the substantial computational and environmental costs of training and deployment. Future research should prioritize lightweight models for edge computing, exploration of novel architectures, and development of benchmarks for causal reasoning. By guiding innovation toward these challenges, the community can unlock the full potential of multimodal geospatial foundation models and advance intelligent, reliable, and responsible Earth observation.


\bibliographystyle{IEEEtran}
\bibliography{reference}

\vspace{-10 mm}
\begin{IEEEbiography}[{\includegraphics[width=1in,height=1.25in,clip,keepaspectratio]{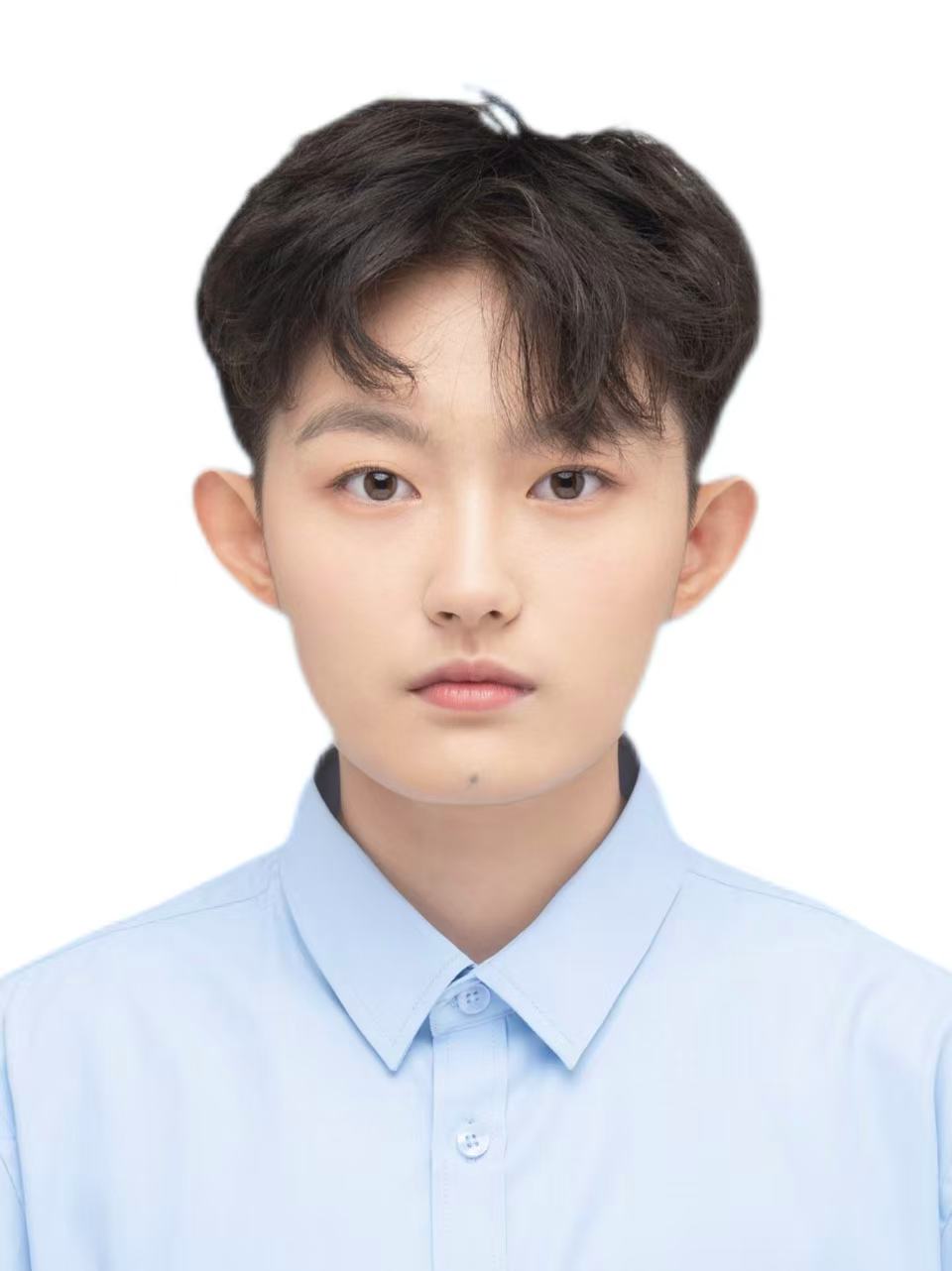}}]{Liling Yang} received the B.E. degree in Robot Engineering from Hunan University of Science and Technology, Xiangtan, China, in 2022, and the M.S. degree in Control Science and Engineering from China University of Geosciences, Wuhan, China, in 2025. She is currently pursuing the Ph.D. degree with the Laboratory of Vision and Image Processing, Hunan University, Changsha, China. Her research interests include computer vision, machine learning, and pattern recognition.
\end{IEEEbiography}

\vspace{-10 mm}
\begin{IEEEbiography}[{\includegraphics[width=1in,height=1.25in,clip,keepaspectratio]{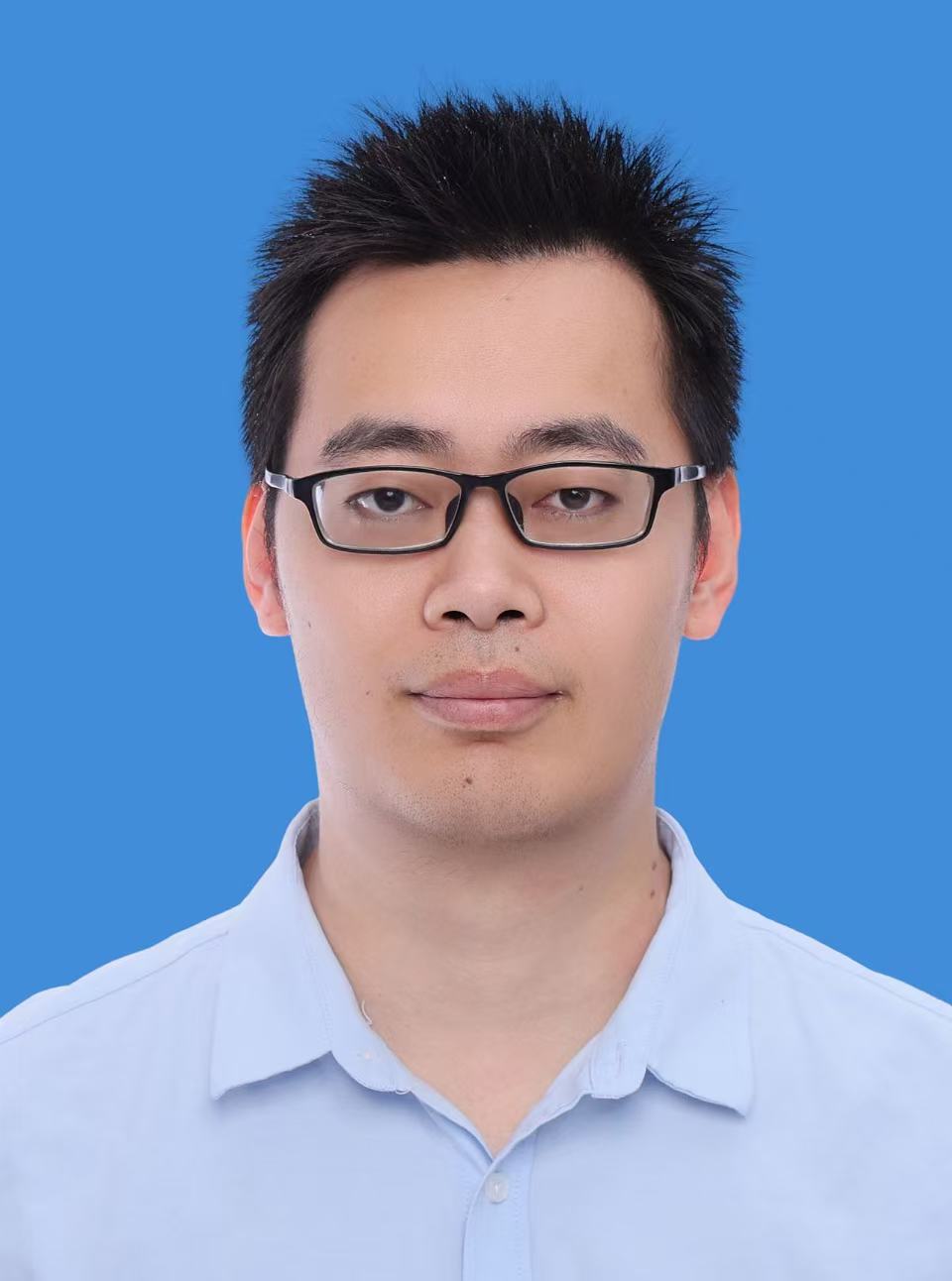}}]{Ning Chen} received the B.S. degree from the School of Earth and Space Sciences, Peking University, Beijing, China, in 2016, and the M.S. degree in GIS from the School of Earth and Space Sciences, Peking University, Beijing, China in 2019. He is currently an Engineer with the Institute of Remote Sensing and Geographic Information System, Peking University. His research interests include satellite image understanding, recommendation system, large-scale sparse learning and pattern recognition. He serves as a reviewer for IEEE Transactions on Geoscience and Remote Sensing, IEEE Geoscience and Remote Sensing Letters, IEEE Sensors Journal, etc. 
\end{IEEEbiography}

\begin{IEEEbiography}[{\includegraphics[width=1in,height=1.25in,clip,keepaspectratio]{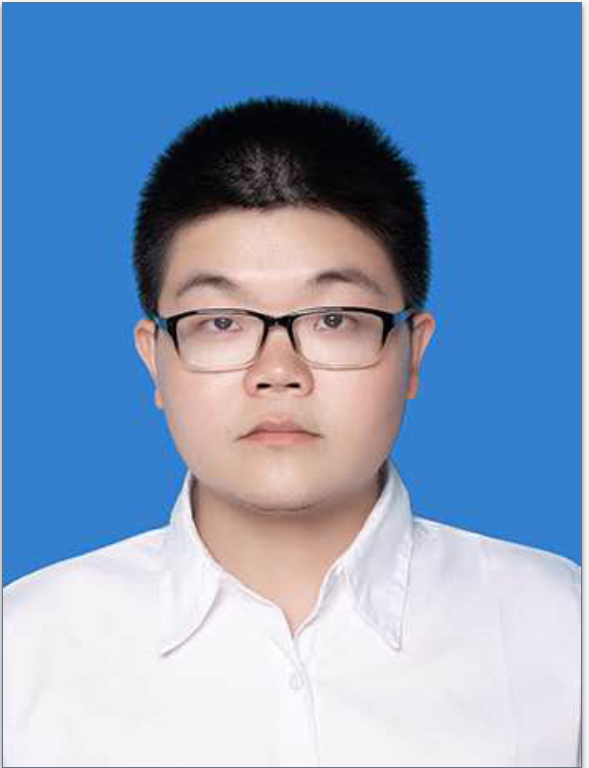}}]{Jun Yue} (Member, IEEE) received the B.Eng. degree in geodesy from Wuhan University, Wuhan, China, in 2013, and the Ph.D. degree in GIS from Peking University, Beijing, China, in 2018.,He is currently an Associate Professor with the School of Automation, Central South University, Changsha, China. His research interests include satellite image analysis, weakly supervised learning, and hyperspectral image interpretation.,Dr. Yue serves as an Associate Editor for Neurocomputing. In addition, he is a reviewer for several prestigious journals, including IEEE Transactions on Pattern Analysis and Machine Intelligence, IEEE Transactions on Image Processing, IEEE Transactions on Geoscience and Remote Sensing, and IEEE Transactions on Neural Networks and Learning Systems.
\end{IEEEbiography}

\vspace{-15 mm}
\begin{IEEEbiography}[{\includegraphics[width=1in,height=1.25in,clip,keepaspectratio]{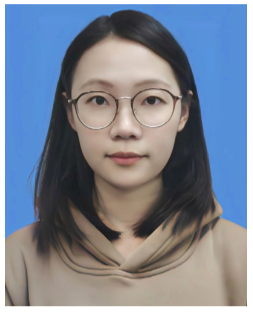}}]{Yidan Liu} received the B.E. and M.S. degrees in telecommunications engineering from Xidian University, Xi'an, China, in 2020 and 2023, respectively. She is currently pursuing the Ph.D. degree with the Laboratory of Vision and Image Processing, Hunan University, Changsha, China. Her research interests include computational imaging, image processing, and deep learning.
\end{IEEEbiography}

\vspace{-15 mm}
\begin{IEEEbiography}[{\includegraphics[width=1in,height=1.25in,clip,keepaspectratio]{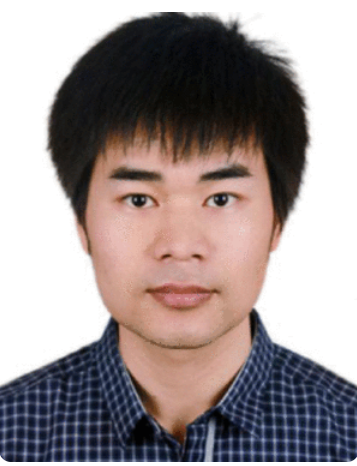}}]{Jiayi Ma} (Senior Member, IEEE) received the B.S. degree in information and computing science and the Ph.D. degree in control science and engineering from the Huazhong University of Science and Technology, Wuhan, China, in 2008 and 2014, respectively. He is currently a Professor with the Electronic Information School, Wuhan University, Wuhan, China. He has coauthored more than 400 refereed journal and conference papers, including Cell, IEEE TPAMI, IJCV, etc. He is a recipient of the Information Fusion Best Paper Award 2024, and the Hsue-shen Tsien Paper Award 2023. He is an Area Editor of Information Fusion, an Associate Editor of IEEE/CAA Journal of Automatica Sinica, Neurocomputing, Geo-spatial Information Science, and Image and Vision Computing, and a Youth Editor of The Innovation and Fundamental Research.
\end{IEEEbiography}

\vspace{-15 mm}
\begin{IEEEbiography}[{\includegraphics[width=1in,height=1.25in,clip,keepaspectratio]{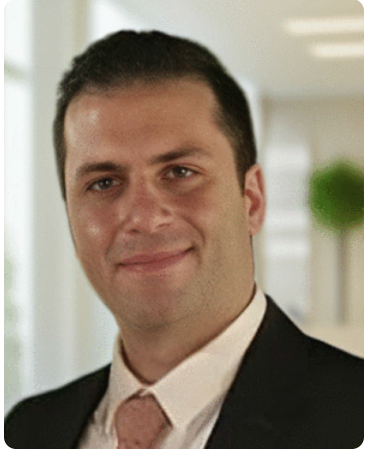}}]{Pedram Ghamisi} (Senior Member, IEEE) received the Ph.D. degree in electrical and computer engineering from the University of Iceland, Reykjavik, Iceland, in 2015.,He is currently the Head of the Machine Learning Group with Helmholtz-Zentrum Dresden-Rossendorf (HZDR), Dresden, Germany, and the Visiting Full Professor with Lancaster University, Lancaster, U.K. Previously, he was a Senior Principal Investigator, Research Professor, and Group Leader of AI4RS with the Institute of Advanced Research in Artificial Intelligence, vienna, Austria. Additionally, he co-founded and served as the CTO of VasogNosis, a startup company based in the U.S. with branches in Milwaukee and California. His research interests include deep learning for remote sensing applications, with particular emphasis on Open Science, AI for Good, Responsible AI, and AI Security, playing a key role in advancing the AI4EO era.,Prof. Ghamisi was the recipient of the 10 distinguished international awards and recognitions, including the IEEE Geoscience and Remote Sensing Society (IEEE GRSS) Highest Impact Paper Award (2020 and 2024), the IEEE GRSS Data Fusion Contest Winner (2017), the IEEE Mikio Takagi Prize (2013), the IEEE GRSS Best Reviewer Prize (2017), and the multiple prestigious scholarships and grants, including the Helmholtz Foundation Models Initiatives (2024), the High Potential Program (HPP) Team Leadership (2018), and the Alexander von Humboldt Fellowship (2015). Since 2021, he has been consistently ranked among the top 1\% of most-cited researchers by Clarivate.
\end{IEEEbiography}

\begin{IEEEbiography}[{\includegraphics[width=1in,height=1.25in,clip,keepaspectratio]{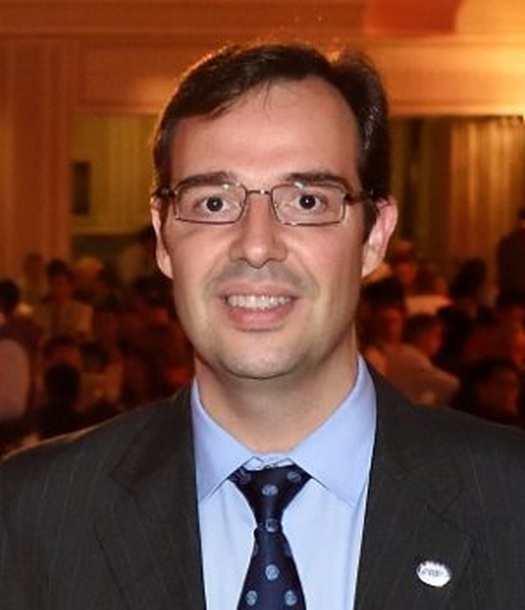}}]{Prof. Antonio Plaza} is a Full Professor and the Head of the Hyperspectral Computing Laboratory at the Department of Technology of Computers and Communications, University of Extremadura, Spain, where he received the M.Sc. degree in 1999 and the PhD degree in 2002, both in Computer Engineering. His main research interests comprise hyperspectral data processing and parallel computing of RS data. He has authored more than 900 publications in this field, including more than 500 JCR journal papers, 25 book chapters, and over 330 peer-reviewed conference proceeding papers. He has guest edited 24 special issues for different journals. Prof. Plaza is a Fellow of IEEE “for contributions to hyperspectral data processing and parallel computing of Earth observation data,” a Fellow of the Asia-Pacific Artificial Intelligence Association (AAIA), and an elected member of Academia Europaea, The Academy of Europe. He is a recipient of the recognition of Best Reviewers of the IEEE Geoscience and Remote Sensing Letters (in 2009) and a recipient of the recognition of Best Reviewers of the IEEE Transactions on Geoscience and Remote Sensing (in 2010), for which he served as Associate Editor in 2007-2012. He was also an Associate Editor for IEEE Access (receiving the recognition of Outstanding Associate Editor in 2017). He is a recipient of the Highest Impact Paper Award of the IEEE Geoscience and Remote Sensing Society (GRSS) in 2024, the Best Column Award of the IEEE Signal Processing Magazine in 2015, the 2013 Best Paper Award of the IEEE Journal of Selected Topics in Applied Earth Observations and Remote Sensing (J-STARS), and the most highly cited paper (2005-2010) in the Journal of Parallel and Distributed Computing. He served as Director of Education Activities for IEEE GRSS in 2011-2012, and President of the Spanish Chapter of IEEE GRSS in 2012-2016. He is currently serving as Chair of the Publications Awards and Fellow Evaluation Committees of IEEE GRSS. He served as Editor-in-Chief of the IEEE Transactions on Geoscience and Remote Sensing journal for five years (2013-2017), and also as Editor-in-Chief of the IEEE Journal on Miniaturization for air and Space Systems (2019-2020). He has been included in the 2018-2024 Highly Cited Researchers List of Clarivate Analytics (seven consecutive years). Additional information: http://sites.google.com/view/antonioplaza.
\end{IEEEbiography}

\begin{IEEEbiography}[{\includegraphics[width=1in,height=1.25in,clip,keepaspectratio]{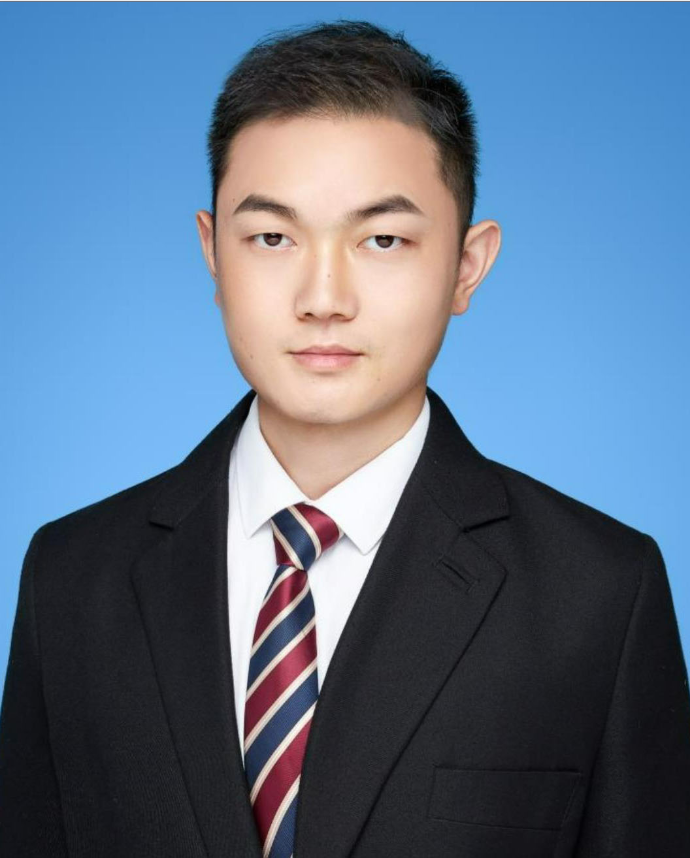}}]{Leyuan Fang}
	(Senior Member, IEEE) received the Ph.D. degree from the College of Electrical and Information Engineering, Hunan University, Changsha, China, in 2015. 
	From August 2016 to September 2017, he was a Postdoc Researcher with the Department of Biomedical Engineering, Duke University, Durham, NC, USA. He is currently a Professor with the School of Artificial Intelligence and Robotics, Hunan University.
	His research interests include sparse representation and multi-resolution analysis in remote sensing and medical image processing. He is the associate editors of IEEE Transactions on Image Processing, IEEE Transactions on Geoscience and Remote Sensing, IEEE Transactions on Neural Networks and Learning Systems, and Neurocomputing. He was a recipient of one 2nd-Grade National Award at the Nature and Science Progress of China in 2019.  
\end{IEEEbiography}

\vfill

\end{document}